\documentclass[twoside,11pt]{article}

\usepackage{jmlr2e}
\usepackage{amsmath}
\usepackage{lipsum}
\usepackage{float}
\usepackage{subcaption}
\allowdisplaybreaks

\newenvironment{nospaceflalign*}
 {\setlength{\abovedisplayskip}{0pt}\setlength{\belowdisplayskip}{2.0pt}%
  \csname flalign*\endcsname}
 {\csname endflalign*\endcsname\ignorespacesafterend}

\firstpageno{1}

\begin{document}

\title{Bounding the Rademacher Complexity\\
of Fourier neural operators}

\author{\name Taeyoung Kim \email legend@snu.ac.kr \\
       \addr Department of Mathematical Science\\
       Seoul National University\\
       Seoul 08826, South Korea
       \AND
       \name Myungjoo Kang \email mkang@snu.ac.kr \\
       \addr Department of Mathematical Science\\
       Seoul National University\\
       Seoul 08826, South Korea}

\maketitle


\abstract{A Fourier neural operator (FNO) is one of the physics-inspired machine learning methods. In particular, it is a neural operator. In recent times, several types of neural operators have been developed, e.g., deep operator networks, Graph neural operator (GNO), and Multiwavelet-based operator (MWTO). Compared with other models, the FNO is computationally efficient and can learn nonlinear operators between function spaces independent of a certain finite basis. In this study, we investigated the bounding of the Rademacher complexity of the FNO based on specific group norms. Using capacity based on these norms, we bound the generalization error of the model. In addition, we investigated the correlation between the empirical generalization error and the proposed capacity of FNO. From the perspective of our result, we inferred that the type of group norms determines the information about the weights and architecture of the FNO model stored in the capacity. And then, we confirmed these inferences through experiments. Based on this fact, we gained insight into the impact of the number of modes used in the FNO model on the generalization error. And we got experimental results that followed our insights.}




\maketitle

\section{Introduction}\label{sec1}

Physics-inspired machine learning is an actively studied area, in which two approaches exist. One approach includes the deep Ritz method (\cite{Weinan:18}), PINNs (\cite{Raissi:19}), and LSNN (\cite{Cai:21}), and the other approach includes
DeepONets (\cite{Lu:21}), MWTO (\cite{Gupta:21}), GNO (\cite{Li:20}), and Fourier neural operator (FNO) (\cite{Li:21}). The former approach focuses on determining solutions to PDEs for fixed PDE and boundary conditions, whereas the latter focuses on
operators between function spaces. In this study, we focus on the FNO, which uses a Fourier transform to quickly and practically manage the convolution operator between two functions. One of the advantages of the FNO is its computational efficiency compared with those of other methods, and unlike DeepONet, its representation is not limited to finite-dimensional space spanned by few basis functions. Previous studies (\cite{Li:21} and \cite{Pathak:22}) confirmed that the FNO can successfully approximate a numerical solver and real-world data, thereby indicating its computational efficiency and potential applicability. Unlike real-world machine-learning problems, approximating the solver operator of the PDE is deterministic and concrete. There is a result of the universal approximation property of the FNO and its approximation error on certain PDE problems (\cite{Kovachki:21}). However, there is no result for estimating the generalization error of the FNO. Although approximating the solver operator of the PDE is deterministic problem, we can provide only a finite number of samples to the FNO. Therefore, the accuracy of inference on hidden data is another problem that needs to be considered.
Several approaches with regard to the bounding generalization error of deep neural networks, e.g., the group norm of weights (\cite{Neysh:15}), spectral norm (\cite{Bart:21}), path norm (\cite{Neysh:15}), Fisher-Rao norm (\cite{Liang:19}), and relative flatness (\cite{Petzka:21}), exist. In this study, we investigate the bounding of generalization errors in the framework of the PAC learning theory. In particular, we bound the Rademacher complexity of the FNO.

\subsection{Overview of FNOs}

\begin{figure}[htp]
\centering
\includegraphics[height=5.5cm]{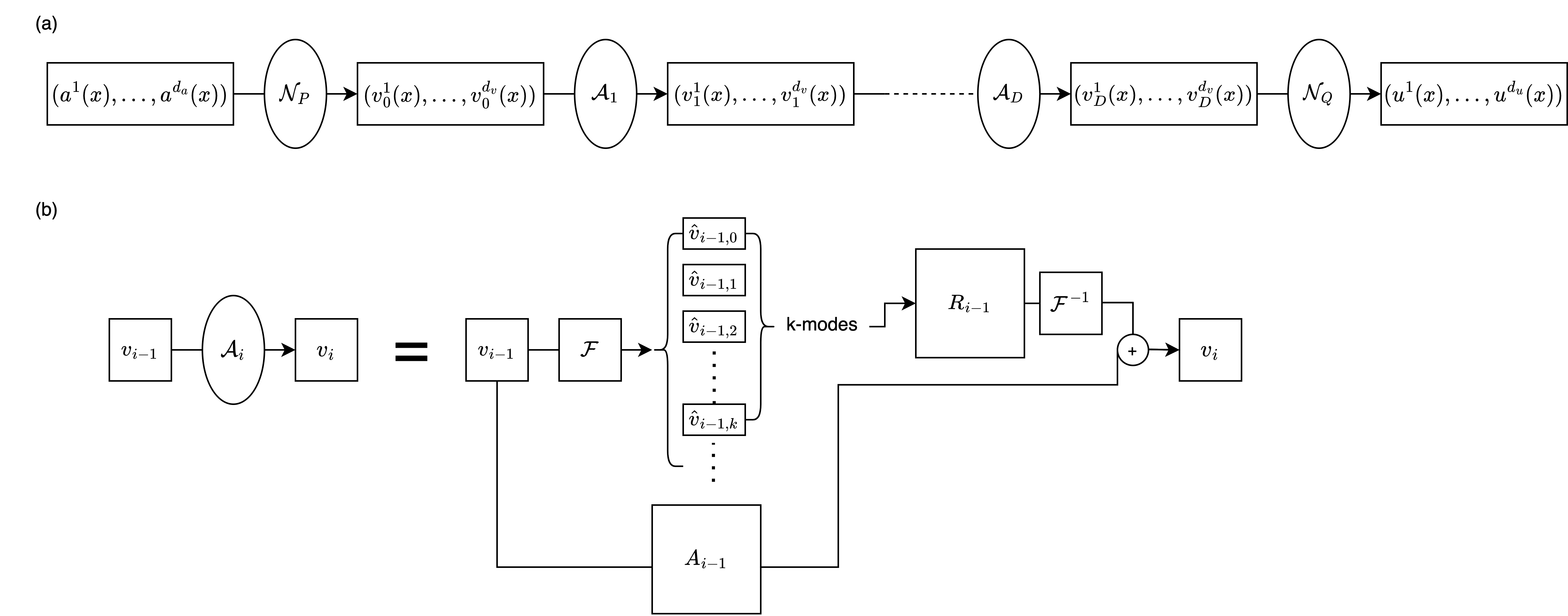}
\caption{(a) Sketch of overall architecture of FNO (b) Detailed diagram of fourier layers}\label{FNOdiagram}
\end{figure}

\noindent
Figure \ref{FNOdiagram} shows the overall structure of the FNO architecture. The input of the network is $\mathbb{R}^{d_{a}}$-valued function on the domain $\tilde{D} \subset \mathbb{R}^{d}$. We denote the input function space of the FNO by $\mathcal{A}(\tilde{D};\mathbb{R}^{d_{a}})$. The vector value of the input function is lifted to the $d_{v}$-dimensional vector using a layer defined as $\mathcal{N}_{P}$. While passing through Fourier layers (which is denoted as $\mathcal{A}_{i}$ in the diagram) iteratively, it is processed as a $\mathbb{R}^{d_{v}}$-valued function. Each Fourier layer comprises the activation function, the sum of a neural network with the convolution of the input function with a kernel parameterized by weight $R_{i}$. After passing through the Fourier layers, the vector value of the $\mathbb{R}^{d_{v}}$-valued function $v_{D}$ is projected onto the $d_{u}$-dimensional vector using $\mathcal{N}_{Q}$. We denote the output function space of the FNO by $\mathcal{U}(\tilde{D};\mathbb{R}^{d_{u}})$. Neural network $A_{i}$ in the Fourier layers can be arbitrarily chosen. In our results, we chose $A_{i}$ as the Fully connected network (FCN) or Convolutional neural network (CNN). As computational machines cannot handle infinite dimensional data, we constructed the FNO model using finite parameters based on the aforementioned concept with regard to real-world implementation.

\subsection{Probably approximately correct (PAC) learning}

PAC learning is a framework of the statistical learning theory proposed by Leslie Valiant in 1984 (\cite{Valiant:84}). One of the main concepts of the PAC learning theory is the no free lunch (NFL) theorem, which states that it is not possible to simultaneously achieve low approximation and estimation errors. The tradeoff between such errors is closely related to the complexity of the hypothesis class. Various quantities related to the complexity of the hypothesis class determine the learnability and decay of estimation errors, e.g., VC dimension, Rademacher complexity, and Gaussian complexity. All the complexities are related; however, there are several differences. For example, the VC dimension is independent of training sets, and the others are not. Neural networks and deep learning, as a subcategory of machine learning, can be applied to the PAC learning theory. In recent times, various studies have been conducted on bounding the Rademacher complexity and VC dimension of the hypothesis class of neural networks. For instance, results with regard to the bounding of Rademacher complexities for FCN (\cite{Neysh:15}), RNN (\cite{Minshuo:20}), GCN (\cite{Lv:21}), and the analysis of the VC dimension of neural networks (\cite{Sontag:98}) have been obtained. In addition, there is information about the bounding Rademacher complexity of DeepONet (\cite{Gopalani:22}) which is also one kind of a neural operator. (\cite{Weinan:20}) estimated the generalization error of ResNet in prior and posterior estimates.

\subsection{Our Contributions}

In this study, we define the capacities of FNO models based on certain types of group norms. And we bound the Rademacher complexity of the hypothesis class based on these capacities for two kinds of FNOs (Fourier layers with FCN and CNN) and induce the bounding of posterior generalization error of the FNO models. In Section 4, we experiment with the data generated from the Burgers equation problem and verify the correlation between our bounding process and empirical generalization errors. And through experiments, we gained insights into the information of model architecture and model weights contained in various types of capacity. We also qualitatively confirmed that empirical generalization errors depend on the number of modes used in the FNO model.

\section{Preliminary}\label{sec2}

\noindent
{\bf Notation} {Several indices have been considered in our discussion. Therefore, to simplify the formulas, we denote $x_{1}\dots x_{d}$ as $\textbf{x}$ and $k_{1}\dots k_{d}$ as $\textbf{k}$. In addition, for the multi-index tensor in the norm, indexes denoted as $\cdot$ are used in the calculation of the norm, e.g.,

\begin{equation*}
\begin{gathered}
\|A_{xy\cdot}\|_{p} = \sqrt[\leftroot{-1}\uproot{4}p]{\sum_{i}{\Big(A_{xyi}\Big)^{p}}}.
\end{gathered}
\end{equation*}

}\

\noindent
{\bf Discretization of data} {
As the function space is infinite-dimensional, to treat the data and operator numerically, we discretize the domain of the function 
and consider the function to be a finite-dimensional vector. Let $\tilde{D}_{N}=\{x_{1},...,x_{N}\}$ be the discretization of domain $\tilde{D}\in\mathbb{R}^{d}$. Then, the $\mathbb{R}^{m}$-valued function $f$ is discretized into $(f(x_{1}),...,f(x_{N})))\in\mathbb{R}^{N\times m}$. Then, we discretize $\mathcal{A}(\tilde{D};\mathbb{R}^{d_{a}})$ and 
$\mathcal{U}(\tilde{D};\mathbb{R}^{d_{u}})$ as $\mathbb{R}^{N\times d_{a}}$ and $\mathbb{R}^{N\times d_{u}}$, respectively. Then, sample data are defined as follows: 
element $((a_{jk}),(u_{jk}))\in\mathbb{R}^{N\times d_{a}}\times\mathbb{R}^{N\times d_{u}}$.

}\

\noindent
{\bf Fourier transform} {
Based on the Fourier analysis, we know that the Fourier transform transfers the convolution operation to pointwise multiplication. For the function of domain $\tilde{D} \subset \mathbb{R}^{d}$, let $\mathcal{F}$ and $\mathcal{F}^{-1}$ be the Fourier and inverse Fourier transforms over $\tilde{D}$, respectively. Thus, we obtain the following relation:
\begin{equation*}
\begin{gathered}
f*k=\mathcal{F}^{-1}(\mathcal{F}(k)\cdot \mathcal{F}(f)).
\end{gathered}
\end{equation*}
For our analysis, we select $\tilde{D}$ as $[0,2\pi]^{d}$. As we treat functions as discretized vectors, we can treat the Fourier transform as a discrete Fourier transform. If the discretization of $\tilde{D}$ is uniform, it can be replaced with a fast Fourier transform. Consider that $\tilde{D}$ is discretized uniformly by resolution $N_{1}\times\dots\times N_{d} = N$, then, for discretized function $f \in \mathbb{R}^{N}$, its FFT $\mathcal{F}(f)(k)$ and IFFT  $\mathcal{F}^{-1}(f)(k)$ are defined as follows:
\begin{equation*}
\begin{gathered}
\mathcal{F}(f)(k)=\frac{1}{\sqrt{N_{1}\dots N_{d}}}\sum_{x_{1}}^{N_{1}}\dots\sum_{x_{d}}^{N_{d}}f(x_{1},...,x_{d})e^{-2i\pi\sum_{j=1}^{d}{\frac{x_{j}k_{j}}{N_{j}}}} \\
\mathcal{F}^{-1}(f)(k)=\frac{1}{\sqrt{N_{1}\dots N_{d}}}\sum_{x_{1}}^{N_{1}}\dots\sum_{x_{d}}^{N_{d}}f(x_{1},...,x_{d})e^{2i\pi\sum_{j=1}^{d}{\frac{x_{j}k_{j}}{N_{j}}}}.
\end{gathered}
\end{equation*}
For our analysis, we denote the components of the FFT and IFFT tensors as \\
$F_{\textbf{k}\textbf{x}}=\frac{1}{\sqrt{N_{1}\dots N_{d}}}e^{-2i\pi\sum_{j=1}^{d}{\frac{x_{j}k_{j}}{N_{j}}}}$, $F^{\dag}_{\textbf{x}\textbf{k}}=\frac{1}{\sqrt{N_{1}\dots N_{d}}}e^{2i\pi\sum_{j=1}^{d}{\frac{x_{j}k_{j}}{N_{j}}}}$, respectively.

}\

\noindent
{\bf Definition 1 (General FNO)} {Let $\tilde{D}_{N}$ be the discretized domain in $\mathbb{R}^{d}$; then, $ \textbf{FNO} : \mathbb{R}^{N\times d_{a}} \rightarrow  \mathbb{R}^{N\times d_{u}}$   is defined as follows:
\begin{equation*}
\begin{gathered}
\textbf{FNO}=\mathcal{N}_{Q}\circ\mathcal{A}_{D}\circ\mathcal{A}_{D-1}\cdots\circ\mathcal{A}_{1}\circ\mathcal{N}_{P},
\end{gathered}
\end{equation*}
where $\mathcal{N}_{P}$ and $\mathcal{N}_{Q}$ denote the neural networks for lifting and projection, respectively. Each $\mathcal{A}_{i}$ is a Fourier layer, For simplicity, we assume that $\mathcal{N}_{Q}$ and $\mathcal{N}_{P}$ are linear maps. Each Fourier layer is a composition of the activation function with a sum of convolutions based on a parameterized function and linear map. Only partial frequencies are used in the Fourier layers. The frequencies used in the model are expressed in an index set $K=\{(k_{1},...,k_{d})\in \mathbb{Z}^{d} : 0 \leq k_{j} \leq k_{max,j}, j=1,...,d\}$. The detailed formula of the FNO is
\begin{flalign*}
v_{0}&:=\mathcal{N}_{P}(a)=\sum_{k}{P_{jk}a_{\textbf{x}k}}  \\
v_{t+1}&:=\mathcal{A}_{t+1}(v_{t})=\sigma\bigg(A_{t+1}v_{t}+\mathcal{F}^{-1}\Big(R_{t+1}\cdot(\mathcal{F}(v_{t}))\Big)\bigg).  \\
&=\sigma\Big(\sum_{\textbf{z},k}{{A_{t+1,\textbf{x}\textbf{z}jk}v_{t,\textbf{z}k}}} + \sum_{\textbf{z},\textbf{k}\in K,k}   {F^{\dag}_{\textbf{x}\textbf{k}}R_{t+1,\textbf{k},jk}F_{\textbf{k}\textbf{z}}v_{t,\textbf{z}k}}\Big)  \quad (t=0,...,D-1) \\
u&:=\sum_{k}{v_{D,\textbf{x}k}Q_{kj}}.
\end{flalign*}

}

\noindent
{\bf CNN layer} {For each Fourier layer, we can replace the general linear map with a CNN layer. A schematic diagram of the convolution with 2D data and a kernel are shown in Figure \ref{CNN}.
\begin{figure}[htp!]
\centering
\includegraphics[height=5.0cm]{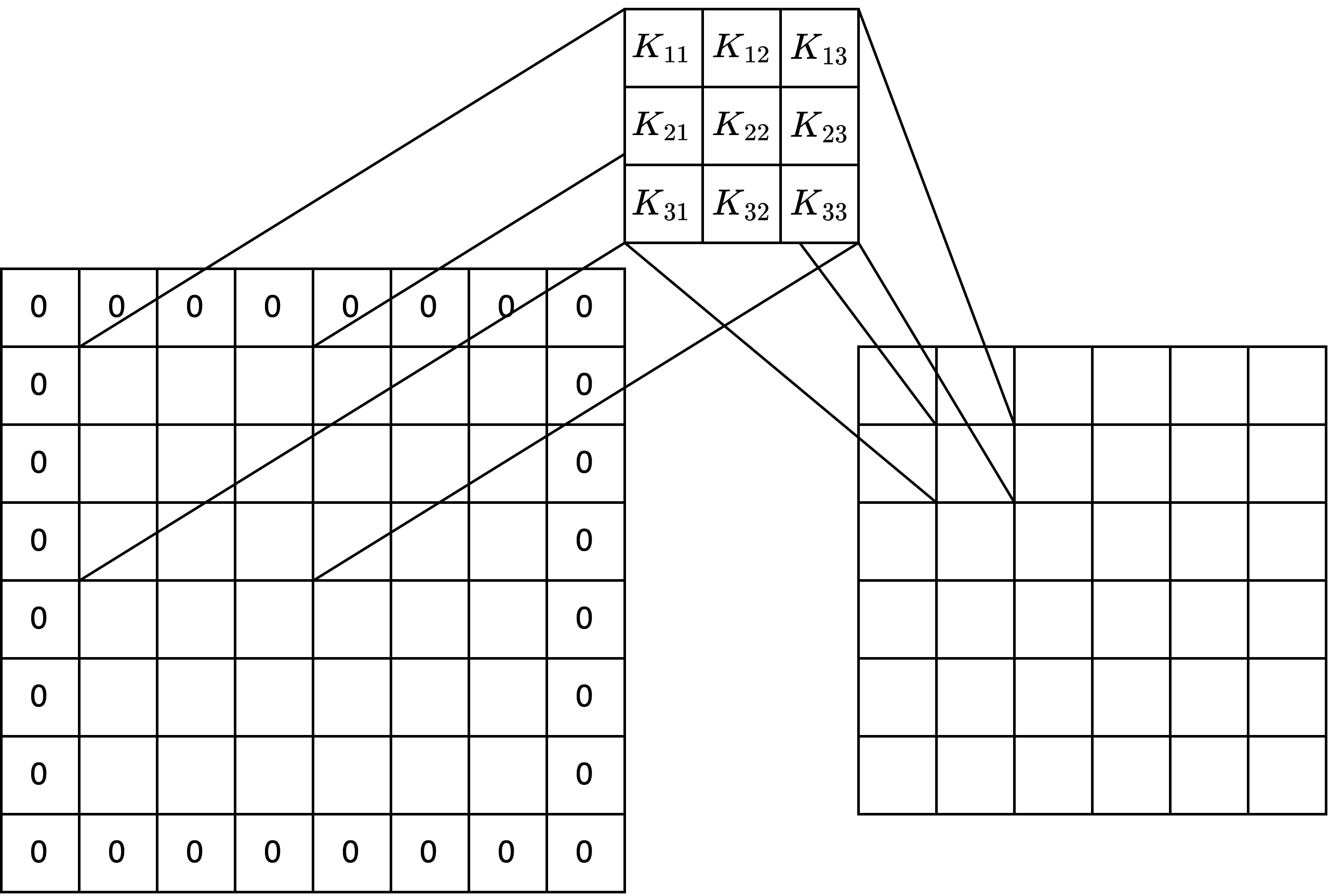}
\caption{schematic diagram of 2D-CNN layer}\label{CNN}
\end{figure}
\noindent
A certain size of kernel swipes the input tensors so that resulting for each index of output an inner product with kernel and local components of the input tensor centering the index. For example, for a $d$-rank input tensor with a size of $N_{1}\times\cdots\times N_{d}$, we consider a $d$-rank tensor kernel $K$ with a size of $c_{1}\times\cdots\times c_{d}$, where each value of $c_{i}$ is less than that of $N_{i}$. Let us denote this CNN layer by the kernel $C(c_{1}\times\cdots\times c_{d})$; then, the tensor that passes through the CNN layer with $K$ is defined as follows:
\begin{equation*}
\begin{gathered}
C(c_{1},\dots,c_{d})(x_{x_{1}\cdots x_{d}})_{z_{1}\cdots z_{d}} = \sum_{j_{1}=0}^{c_{1}-1}\cdots\sum_{j_{d}=0}^{c_{d}-1} K_{j_{1},\dots,j_{d}}x_{z_{1}+j_{1},\dots,z_{d}+j_{d}}.
\end{gathered}
\end{equation*}
As the positional dimensions of the tensor must be maintained, the CNN layers in our study are restricted to kernels of odd sizes. To fit the dimensions, padding is applied to the input tensor of the CNN layer. For example, for $N_{1}\times\cdots\times N_{d}$-dimensional tensor $x_{x_{1}\cdots x_{d}}$ and the CNN layer $C(c_{1},\dots,c_{d})$, we pad $\frac{c_{i}-1}{2}$ zeros for each side of the input tensor. We denote this padded tensor by $\tilde{x}$. Then, $C(c_{1},\dots,c_{d})(\tilde{x}_{x_{1}\cdots x_{d}})$ has the same dimension as the input tensor.
As the number of channels in the Fourier layers is fixed, for a CNN layer with multiple channels, we use the same notation, i.e., $C(c_{1},\dots,c_{d})$, and the detailed formula for such multi-channel CNN layer can be defined as follows:
\begin{equation*}
\begin{gathered}
C(c_{1},\dots,c_{d})(x_{x_{1}\cdots x_{d}})_{z_{1}\cdots z_{d}j} = \sum_{k=1}^{d_{u}}\sum_{j_{1}=0}^{c_{1}-1}\cdots\sum_{j_{d}=0}^{c_{d}-1} K_{j_{1},\dots,j_{d},k,j}x_{z_{1}+j_{1},\dots,z_{d}+j_{d},k}.
\end{gathered}
\end{equation*}

}\

\noindent
{\bf Definition 2 (FNO with the CNN layer)} {Consider the settings of the aforementioned FNO; the only difference is that the Fourier layer is a sum of the CNN layer and convolution with parameterized functions.
\begin{flalign*}
v_{t+1}&:=\mathcal{A}_{t+1}(v_{t})=\sigma\bigg(C_{t+1}(c_{1},\dots,c_{d})(\tilde{v_{t}})+\mathcal{F}^{-1}\Big(R_{t+1}\cdot(\mathcal{F}(v_{t}))\Big)\bigg)  \\
&=\sigma\Big(\sum_{k=1}^{d_{u}}\sum_{j_{1}=0}^{c_{1}-1}\cdots\sum_{j_{d}=0}^{c_{d}-1} K_{t+1,jk,j_{1},\dots,j_{d}}\tilde{v_{t}}_{x_{1}+j_{1},\dots,x_{d}+j_{d},k} \\
&+\sum_{\textbf{z},\textbf{k}\in K,k}{{F_{\textbf{x}\textbf{k}}}^{\dag}R_{t+1,\textbf{k},jk}F_{\textbf{k}\textbf{z}}v_{t,\textbf{z}k}}\Big). 
\end{flalign*}
} \\
\noindent
An ideal operator should infer solution from all the functions in the input function space. But for practical and implemental reasons, finite training samples are selected from 
distributions on the vector space, which is a discretized function space. Suppose $\mathcal{D}$ is a distribution on $\mathbb{R}^{N\times d_{a}}\times\mathbb{R}^{N\times d_{u}}$; then, we define the loss functions as follows: 
\\
\\
\noindent
{\bf Definition 3 (Loss for FNO)} {Suppose that the training dataset is given by
\begin{equation*}
\begin{gathered}
S:=\{((a_{i,jk}),(u_{i,jk}))\in\mathbb{R}^{N\times d_{a}}\times\mathbb{R}^{N\times d_{u}}:i=1,...,m\},
\end{gathered}
\end{equation*}
where each sample is independently chosen from the distribution $\mathcal{D}$; 
the training loss is defined as follows:
\begin{equation*}
\begin{gathered}
\mathcal{L}_{S}(a):=\frac{1}{m}\sum_{i=1}^{m}\Big(u_{i,jk}-\textbf{FNO}(a_{i,jk})\Big)^{2}.
\end{gathered}
\end{equation*}
Let $p$ be the probability distribution of $\mathcal{D}$, which is defined as $\mathbb{R}^{N\times d_{a}}\times\mathbb{R}^{N\times d_{u}}$. Then, 
the loss of the entire distribution $\mathcal{D}$ is defined as follows:
\begin{equation*}
\begin{gathered}
\mathcal{L}_{\mathcal{D}}(a):=\int_{\mathbb{R}^{N\times d_{a}}\times\mathbb{R}^{N\times d_{u}}}{\Big(u_{i,jk}-\textbf{FNO}(a_{i,jk})\Big)^{2}}dp.
\end{gathered}
\end{equation*}

}

\section{Generalization bound for FNOs}\label{sec3}
In this section, we calculate the upper bound of the Rademacher complexity of the FNO; based on this bounding, we estimate the generalization bound. We demonstrate several lemmas concerning our main results. The proof of the main theorems comprises two main lemmas: inequality for the Rademacher complexity part and supremum of the norm of FNO models. Using these lemmas, we prove our main results. 

\subsection{Mathematical Setup}
\noindent
{\bf Definition 4 (Rademacher complexity)} {Let $\mathcal{F}$ be a class of mapping from $\mathcal{X}$ to $\mathbb{R}$. Suppose $\{x_{i}\in\mathcal{X}:i=1,...,m\}$ is given. $\epsilon_{i}$ are independent, uniform, $\{+1,-1\}$-valued random variables. The empirical Rademacher complexity of $\mathcal{F}$ on the given sample set is defined as follows:
\begin{equation*}
\begin{gathered}
\mathcal{R}_{m}(\mathcal{F})=\mathbb{E}_{\epsilon}\bigg[\frac{1}{m}\sup_{f\in \mathcal{F}}\sum_{i=1}^{m}\epsilon_{i}f(x_{i})\bigg].
\end{gathered}
\end{equation*}

}

\noindent
\\
The following definitions are the main components of our results: 
\\
\noindent
{\bf Definition 5 (weight norms and capacity) } {For the multi-rank tensor $M_{i_{1},...,i_{m},j_{1},...,j_{k}}$, we define the following weight norm:
\begin{equation*}
\begin{gathered}
\|M_{i_{1},...,i_{m},j_{1},...,j_{k}}\|_{p:\{i_{1},...,i_{m}\},q:\{j_{1},...,j_{k}\}}:=\sqrt[\leftroot{-1}\uproot{4}q]{\sum_{j_{1}...j_{m}}{\bigg(\sqrt[\leftroot{-1}\uproot{4}p]{\sum_{i_{1}...i_{k}}{M_{i_{1},...,i_{m},j_{1},...,j_{k}}}^{p}}\bigg)^{q}}}.
\end{gathered}
\end{equation*}
For $p=\infty$ or $q=\infty$ cases, we think sup-norm instead of above definition.
Now, suppose for an FNO with a Fourier layer of depth $D$, we denote $Q$ for the weight matrix of projection, $P$ for the weight matrix of lifting, and $A_{i}$ and $R_{i}$ for the weight tensors of the
Fourier layers. Then, we define $\|\cdot\|_{p,q}$, where $p$ is the index for positions, frequencies, and inputs, and $q$ is the index of output. We define the following norm for the Fourier layer:
\begin{equation*}
\begin{gathered}
\|(A_{i},R_{i})\|_{p,q} := \|A_{i}\|_{p,q} + \|R_{i}\|_{p,q}\frac{\sqrt[\leftroot{-1}\uproot{4}p*]{k_{max,1}...k_{max,d}}}{N^{  \lfloor \frac{1}{p*}-\frac{1}{q} \rfloor_{+} }}.
\end{gathered}
\end{equation*}
The capacity of the FNO model $h$ as a product of the weights of its layers is defined as follows:
\begin{equation*}
\begin{gathered}
\gamma_{p,q}(h):=\|P\|_{p,q}\|Q\|_{p,q}\prod_{i=1}^{D}\|(A_{i},R_{i})\|_{p,q}.
\end{gathered}
\end{equation*}
Now, we define the norms for the CNN layers. For the kernel tensor $K$ of the CNN layer, we define the following norms of the weights and capacities of the entire neural network. In the $\|\cdot\|_{p,q}$ norm for the kernel tensor of the CNN layer, $p$ is the index of kernels and input, and $q$ is the index of the output.
\begin{equation*}
\begin{gathered}
\|(K_{i},R_{i})\|_{p,q}:=\|K\|_{p,q}\sqrt[\leftroot{-1}\uproot{4}p*]{c_{1}\dots c_{d}} + \sqrt[\leftroot{-1}\uproot{4}p*]{k_{max,1}...k_{max,d}}\|R\|_{p,q}  \\
{\gamma_{CNN}}_{p,q}(h_{CNN}):=\|P\|_{p,q}\|Q\|_{p,q}\|\prod_{i=1}^{D}\|(K_{i},R_{i})\|_{p,q}.
\end{gathered}
\end{equation*}
}
\noindent
Next, we define hypothesis classes, of which the Rademacher complexity is bounded in our results. A hypothesis class is a collection of functions, from which a learning algorithm selects a function.
\\
\\
\noindent
{\bf Definition 6 (Hypothesis classes of FNO)} {Suppose that the function classes of the FNO with the $D$ depth and maximal modes of Fourier layers are $k_{max,1},...,k_{max,d}$. The width, size of the input vector, size of the output vector, and activation function are fixed. We define the hypothesis class for a general FNO as follows:
\begin{equation*}
\begin{gathered}
\mathcal{H}_{C_{P},C_{0},...,C_{D},C_{Q}}^{d_{in}}:=\{\textbf{FNO}:\|P\|_{p,q}\leq C_{P},\\
\|(A_{i},R_{i})\|_{p,q} \leq C_{i} (i=1,...,D) ,\|Q\|_{p,\infty} \leq C_{Q} \}.
\end{gathered}
\end{equation*}
Finally, we define the hypothesis class of the FNO with the CNN layers as follows:

\begin{equation*}
\begin{gathered}
{\mathcal{H}_{CNN}}_{C_{P},C_{0},...,C_{L},C_{Q}}^{d_{in}}:=\{\textbf{FNO}:\|P\|_{p,q}\leq C_{P},\\
\|(K_{i},R_{i})\|_{p,q} \leq C_{i} (i=1,...,D) ,
\|Q\|_{p,\infty} \leq C_{Q} \}.
\end{gathered}
\end{equation*}
\\We also define the following auxiliary definition for the hypothesis class of sub-neural networks of FNO models, where the terminal layer is the Fourier layer (denoted as $\textbf{FNO}_{sub:i}$).
\begin{equation*}
\begin{gathered}
\mathcal{H}_{C_{P},C_{0},...,C_{i}}^{d_{in}}:=\{\textbf{FNO}_{sub:i}:\|P\|_{p,q}\leq C_{P},
\|(A_{t},R_{t})\|_{p,q} \leq C_{t},(t=1,...,i) 
\}.
\end{gathered}
\end{equation*}
Similarly, we define ${\mathcal{H}_{CNN}}_{C_{P},C_{0},...,C_{i}}^{d_{in}}$.
}

\subsection{Main Results}
Notations in each lemma and theorem are based on the definitions of Section 3.1, and the activation function is Lipschitz continuous. And we set our notations as follows: for a given sample $S=\{a_{i}\}_{i=1,\dots,m}$ (where $a_{i}$ are the input data) and hypothesis class $\mathcal{H}_{C_{P},C_{0},...,C_{t}}^{d_{in}}$, we denote $h(a_{i})$ by $v_{t,i}$ where $h\in\mathcal{H}_{C_{P},C_{0},...,C_{t}}^{d_{in}}$. The components are denoted by $v_{t,i,\textbf{x}j}$.\\

\noindent
The following lemma regarding $l_{p}$ norms is frequently used in our proofs: \\
\\
\noindent
{\bf Lemma 1 (norm inequality)} {If $1\leq p\leq q \leq \infty$, for $v \in \mathbb{R}^N$ we obtain the following inequality:  \\
\begin{equation*}
\begin{gathered}
\|v\|_{q}\leq \|v\|_{p} \leq \|v\|_{q}N^{\frac{1}{p}-\frac{1}{q}} .
\end{gathered}
\end{equation*}
\\
Let $\lfloor \cdot \rfloor_{+}$ denote the ReLU function. Then, for an arbitrary $1\leq p,q$, inequality can be defined as
\begin{equation*}
\begin{gathered}
\|v\|_{p} \leq   \|v\|_{q}N^{\lfloor\frac{1}{p}-\frac{1}{q}\rfloor_{+}}.
\end{gathered}
\end{equation*}\\
}
\noindent
The following lemma is required to handle nonlinear loss in our proof (a proof of this lemma can be found in (\cite{Maurer:16})): \\
\\
\noindent
{\bf Lemma 2 (Vector-contraction inequality for the Rademacher complexity)} {
Assume that $\sigma$ is a Lipschitz continuous function with Lipschitz constant $L$, and $\mathcal{F}$ is a hypothesis class of $\mathbb{R}^{N}$-valued functions. Then we have the following inequality.  \\
\begin{equation*}
\begin{gathered}
\mathbb{E}_{\epsilon}\bigg[\frac{1}{m}\sup_{f\in \mathcal{F}}\sum_{i=1}^{m}\epsilon_{i}\sigma(f(x_{i}))\bigg]\leq
\sqrt{2}L\mathbb{E}_{\epsilon}\bigg[\frac{1}{m}\sup_{f\in \mathcal{F}}\sum_{i,k}\epsilon_{ik}f_{k}(x_{i})\bigg].
\end{gathered}
\end{equation*}\\
} 
\noindent
We now prove our main results. Our proof is composed of two parts. Firstly, we get an upper bound of $p*$-norm of the output of FNO models. And we bound the Rademacher complexity of the FNO model on samples based on the upper bound we found. In our discussion, we assume that the projection and lifting layers are linear maps; however, we can easily generalize this to a general FCN.\\
\\
\noindent
Lemma 3 and 3' are the main factors of our result, in which Fourier layers are inductively peeled off. \\
\\
\noindent
{\bf Lemma 3} {
Suppose $\mathcal{H} = \mathcal{H}_{C_{P},C_{1},...,C_{D},C_{Q}}^{d_{in}}$ is the hypothesis class of the FNO with constants $C_{P},C_{1},\dots,C_{D},C_{Q}$. Then, for a sample $a \in \mathbb{R}^{N\times d_{a}}$, we obtain the following inequality:
\begin{equation*}
\begin{gathered}
\sup_{h \in \mathcal{H}}\|h(a)_{\cdot\cdot}\|_{p*,\infty} \\[5pt] 
\leq  L^{D}(NH)^{D\lfloor\frac{1}{p*}-\frac{1}{q} \rfloor_{+}}H^{\lfloor\frac{1}{p*}-\frac{1}{q}\rfloor_{+}}C_{Q}C_{D}\dots C_{1}C_{P}\|a\|_{p*}
\end{gathered}
\end{equation*}
} 
\begin{proof}

\begin{flalign*}
&h(a)_{\textbf{x}j} \\
&=\sum_{k}v_{D,\textbf{x}k}Q_{kj} \\
&\leq \|v_{D,\textbf{x}\cdot}\|_{p*}\|Q_{\cdot j}\|_{p} \tag{1} \label{eq:1}
\end{flalign*}
Then we have following:
\begin{flalign*}
&\|h(a)_{\cdot\cdot}\|_{p*,\infty} \\
&\leq \sup_{j} \sqrt[\leftroot{-1}\uproot{4}p*]{\sum_{\textbf{x}}\|v_{D,\textbf{x}\cdot}\|^{p*}_{p*}\|Q_{\cdot j}\|^{p*}_{p}}  \\
&\leq \|v_{D,\cdot\cdot}\|_{p*}C_{Q}
\end{flalign*}
Now, we peel off fourier layers.
\begin{flalign*}
&\sigma\Big(A_{D}(a)+\mathcal{F}^{-1}(R_{D}\cdot(\mathcal{F}(a)))\Big)_{\textbf{x}j}\\
&=\sigma\Big(\sum_{\textbf{z},k}{A_{D,\textbf{x}\textbf{z}kj}v_{D-1,\textbf{z}k}}+  \sum_{\textbf{k},\textbf{z},k}{F^{\dag}_{\textbf{x}\textbf{k}}R_{D,\textbf{k},jk}F_{\textbf{k}\textbf{z}}v_{D-1\textbf{z}k}}\Big)\\
&\leq L\Big|\sum_{\textbf{z},k}{A_{D,\textbf{x}\textbf{z}kj}v_{D-1,\textbf{z}k}} + \sum_{\textbf{k},\textbf{z},k}{{F^{\dag}_{\textbf{x}\textbf{k}}R_{D,\textbf{k},jk}F_{\textbf{k}\textbf{z}}v_{D-1,\textbf{z}k}}}\Big| \\
&\leq L\bigg(\|A_{D,\textbf{x}\cdot\cdot j}\|_{p}+ \Big\|\sum F^{\dag}_{\textbf{x}\textbf{k}}R_{D,\textbf{k},j\cdot}F_{\textbf{k}\cdot }\Big\|_{p}\bigg)\Big\|v_{D-1,\cdot\cdot}\Big\|_{p*}, \tag{2} \label{eq:2}  
\end{flalign*}
For $\Big\|\sum F^{\dag}_{\textbf{x}\textbf{k}}R_{D,\textbf{k},j\cdot}F_{\textbf{k}\cdot }\Big\|_{p}$ in \eqref{eq:2},
\begin{flalign*}
\Big\|\sum_{\textbf{k}} F^{\dag}_{\textbf{x}\textbf{k}}R_{D,\textbf{k},j\cdot}F_{\textbf{k}\cdot}\Big\|_{p}  
=\sqrt[\leftroot{-1}\uproot{4}p]{\sum_{\textbf{z},k}{\Big(\sum_{\textbf{k}} F^{\dag}_{\textbf{x}\textbf{k}}R_{D,\textbf{k},jk}F_{\textbf{k}\textbf{z}}\Big)^{p}}}.
\end{flalign*}
For fixed $\textbf{x}, \textbf{z}, k$, $\Big(F^{\dag}_{\textbf{x}\textbf{k}}F_{\textbf{k}\textbf{z}}\Big)_{\textbf{k}}$ is a $k_{max,1},\dots,k_{max,d}$-dimensional vector, where each component exhibits the $\frac{e^{ib}}{N}$ form. Thus, by applying H{\"o}lder{'}s inequality, we obtain the following inequality:
\begin{flalign*}
&\Big\|\sum_{\textbf{k}} F^{\dag}_{\textbf{x}\textbf{k}}R_{D,\textbf{k},j\cdot}F_{\textbf{k}\cdot}\Big\|_{p}  \\
&\leq \sqrt[\leftroot{-1}\uproot{4}p]{\sum_{\textbf{z},k}{\Big({\frac{\sqrt[\leftroot{-1}\uproot{4}p*]{k_{max,1}...k_{max,d}}}{N}\|R_{D,\cdot,jk}\|_{p}\Big)^{p}}}} \\
&= \frac{\sqrt[\leftroot{-1}\uproot{4}p*]{k_{max,1}...k_{max,d}}}{N}\sqrt[\leftroot{-1}\uproot{4}p]{N\sum_{\textbf{k},k}{R_{D,\textbf{k},jk}^{p}}} \\
&= \sqrt[\leftroot{-1}\uproot{4}p*]{\frac{k_{max,1}...k_{max,d}}{N}}\|R_{D,\cdot,j,\cdot}\|_{p}.
\end{flalign*}
Then, we obtain the following bound:
\begin{flalign*}
&\sigma\Big(A_{D}(a)+\mathcal{F}^{-1}(R_{D}\cdot(\mathcal{F}(a)))\Big)_{\textbf{x}j}\\
&\leq  L\Big(\|A_{D,\textbf{x}\cdot\cdot j}\|_{p}+\sqrt[\leftroot{-1}\uproot{4}p*]{\frac{k_{max,1}...k_{max,d}}{N}}\|R_{D,\cdot,j,\cdot}\|_{p}\Big)
\Big\|v_{D-1,\cdot\cdot}\Big\|_{p*}.
\end{flalign*}
So applying the above bound iteratively, we get the following inequality:
\begin{flalign*}
&\sup_{h \in \mathcal{H}_{C_{P},C_{0},...,C_{D}}}\|v_{D,\cdot\cdot}\|_{p*} \\
&\leq \sup_{h \in \mathcal{H}_{C_{P},C_{1},...,C_{D}}}L\Big(\|A_{D}\|_{p,p*} + \sqrt[\leftroot{-1}\uproot{4}p*]{k_{max,1}...k_{max,d}} \|R_{D}\|_{p,p*}\Big) \Big\|v_{D-1,\cdot\cdot}\Big\|_{p*}\\
&\leq (NH)^{\lfloor \frac{1}{p*}-\frac{1}{q} \rfloor_{+}} \sup_{h \in \mathcal{H}_{C_{P},C_{1},...,C_{D}}} \\
&L\Big(\|A_{D}\|_{p,q} + \frac{\sqrt[\leftroot{-1}\uproot{4}p*]{k_{max,1}...k_{max,d}}}{N^{ \lfloor \frac{1}{p*}-\frac{1}{q} \rfloor_{+}}} \|R_{D}\|_{p,q}\Big) \Big\|v_{D-1,\cdot\cdot}\Big\|_{p*}  \tag{3} \label{eq:3}  \\
&\leq L(NH)^{\lfloor \frac{1}{p*}-\frac{1}{q} \rfloor_{+}}C_{D}\sup_{h \in \mathcal{H}_{C_{P},C_{1},...,C_{D-1}}}\Big\|v_{D-1,\cdot\cdot}\Big\|_{p*} \\
&\leq ... \\
&\leq L^{D}(NH)^{D\lfloor\frac{1}{p*}-\frac{1}{q} \rfloor_{+}}C_{D}\dots C_{1}\sup_{h \in \mathcal{H}_{C_{P}}}\Big\|v_{1,\cdot\cdot}\Big\|_{p*}  \\
&\leq L^{D}(NH)^{D\lfloor\frac{1}{p*}-\frac{1}{q} \rfloor_{+}}H^{\lfloor\frac{1}{p*}-\frac{1}{q}\rfloor_{+}}C_{D}\dots C_{1}C_{P}\|a\|_{p*}  \tag{4} \label{eq:4}.
\end{flalign*}
So, combining the two inequalities we got, we have the following inequality:
\begin{flalign*}
&\|h(a)_{\cdot\cdot}\|_{p*,\infty} \\
&\leq \|v_{D,\cdot\cdot}\|_{p*}C_{Q}  \\
&\leq L^{D}(NH)^{D\lfloor\frac{1}{p*}-\frac{1}{q} \rfloor_{+}}H^{\lfloor\frac{1}{p*}-\frac{1}{q}\rfloor_{+}}C_{Q}C_{D}\dots C_{1}C_{P}\|a\|_{p*}
\end{flalign*}
H{\"o}lder{'}s inequality is used in \eqref{eq:1} and \eqref{eq:2}.
And, we used norm inequality in \eqref{eq:3} and \eqref{eq:4}.
\end{proof}
\noindent
\\
The proof of the following lemma is similar to Lemma 3. However, in this case, the hypothesis class is composed of FNO with CNN layers. \\
\\
\noindent
{\bf Lemma 3'} {
Suppose $\mathcal{H} = {\mathcal{H}_{CNN}}_{C_{P},C_{1},...,C_{D},C_{Q}}^{d_{in}}$ is the hypothesis class of an FNO with CNN layer with constants $C_{P},C_{1},\dots,C_{D},C_{Q}$. Then, for a sample $a\in \mathbb{R}^{N\times d_{a}}$, we obtain the following inequality:

\begin{equation*}
\begin{gathered}
\sup_{h \in \mathcal{H}}\|h(a)_{\cdot\cdot}\|_{p*,\infty}  \\[5pt] 
\leq  L^{D}H^{(D+1)\lfloor\frac{1}{p*}-\frac{1}{q} \rfloor_{+}}C_{Q}C_{D}\dots C_{1}C_{P}\|a\|_{p*}
\end{gathered}
\end{equation*}

\begin{proof}
We just need to modify the induction parts of the Fourier layers in the proof of Lemma 3.
\begin{flalign*}
&\sigma\Big(C_{D}(a)+\mathcal{F}^{-1}(R_{D}\cdot(\mathcal{F}(a)))\Big)_{\textbf{x}j}\\
&=\sigma\Big(\sum_{j_{1}=0}^{c_{1}-1}\cdots\sum_{j_{d}=0}^{c_{d}-1}\sum_{k=1}^{d_{v}}K_{D,jk,j_{1},\dots,j_{d}} v_{D-1,x_{1}+j_{1},\dots,x_{d}+j_{d}k} \\
&+\sum{F^{\dag}_{\textbf{x}\textbf{k}}R_{D,\textbf{k},jk}F_{\textbf{k}z_{1}...z_{d}}v_{D-1,i,z_{1}...z_{d}k}}\Big) \\
&\leq  L\bigg(\|K_{D,j,\cdots}\|_{p}\bigg\|v_{D-1,x_{1}+\cdot,\dots,x_{d}+\cdot,\cdot}\bigg\|_{p*} \tag{5} \label{eq:5} \\
&+\sqrt[\leftroot{-1}\uproot{4}p*]{\frac{k_{max,1}...k_{max,d}}{N}}\|R_{D,\cdot,j,\cdot}\|_{p}\bigg\|v_{D-1,\cdot \cdot }\bigg\|_{p*}\bigg).
\end{flalign*}
where we use H{\"o}lder{'}s inequality in \eqref{eq:5}. Then, by applying the $p*$ norm to the aforementioned inequality over $\textbf{x},j$ and the norm inequality, we obtain the following inequality:
\begin{flalign*}
&\bigg\|\sigma\Big(C_{D}(a)+\mathcal{F}^{-1}(R_{D}\cdot(\mathcal{F}(a)))\Big)_{\cdot\cdot}\bigg\|_{p*} \\
&\leq L\bigg(\sqrt[\leftroot{-1}\uproot{4}p*]{\sum_{j}\|K_{D,j,\cdots}\|_{p}^{p*}\sum_{\textbf{x}}\sum_{j_{1}=0}^{c_{1}-1}\cdots\sum_{j_{d}=0}^{c_{d}-1}\sum_{k=1}^{d_{v}}\|v_{D-1,x_{1}+\cdot,\dots,x_{d}+\cdot,\cdot}\|^{p*}
}\\
&+\sqrt[\leftroot{-1}\uproot{4}p*]{k_{max,1}...k_{max,d}}\|R_{D}\|_{p,q} H^{\lfloor \frac{1}{p*}-\frac{1}{q} \rfloor_{+}}\bigg\|v_{D-1,\cdot \cdot }\bigg\|_{p*}   \bigg) \\
&\leq LH^{\lfloor \frac{1}{p*}-\frac{1}{q} \rfloor_{+}}\bigg(\sqrt[\leftroot{-1}\uproot{4}p*]{c_
{1}\dots c_{d}}\|K_{D}\|_{p,q} +\sqrt[\leftroot{-1}\uproot{4}p*]{k_{max,1}\dots k_{max,d}}\|R_{D}\|_{p,q} \bigg) \bigg\|v_{D-1,\cdot \cdot }\bigg\|_{p*}.
\end{flalign*}
The remainder of the proof is similar to that of Lemma 3. 
\end{proof}

\noindent
{\bf Lemma 4} {
Suppose $\mathcal{H}_{C_{P},C_{1},...,C_{D},C_{Q}}^{d_{in}}$ is the hypothesis class of the FNO with given constants $C_{P},C_{1},\dots,C_{D}, C_{Q}$. Then, for samples $S=\{a_{i}\}_{i=1,\dots,m}$, we obtain the following inequality:
\begin{equation*}
\begin{gathered}
\mathbb{E}_{\epsilon}\bigg[\frac{1}{m}\sup_{h \in \mathcal{H}_{C_{P},C_{1},...,C_{D},C_{Q}}^{d_{in}}}\sum_{i,\textbf{x},j}\epsilon_{i\textbf{x}j}h(a_{i})_{\textbf{x}j}\bigg]
\leq \frac{N^{\frac{1}{p}}d_{u}}{m}\sum_{i} \sup_{h \in \mathcal{H}_{C_{P},C_{1},...,C_{D},C_{Q}}^{d_{in}}} \|h(a_{i})_{\cdot\cdot}\|_{p*,\infty}.  
\end{gathered}
\end{equation*}
} 
\begin{proof}
\begin{flalign*}
&\mathbb{E}_{\epsilon}\bigg[\frac{1}{m}\sup_{h \in \mathcal{H}_{C_{P},C_{1},...,C_{D},C_{Q}}^{d_{in}}}\sum_{i,\textbf{x},j}\epsilon_{i\textbf{x}j}h(a_{i})_{\textbf{x}j}\bigg] \\
&\leq \mathbb{E}_{\epsilon}\bigg[\frac{1}{m}\sup_{h \in \mathcal{H}_{C_{P},C_{1},...,C_{D},C_{Q}}^{d_{in}}}\sum_{i,\textbf{x},j}\Big\vert h(a_{i})_{\textbf{x}j}\Big\vert\bigg] \\
&\leq \frac{N^{\frac{1}{p}}d_{u}}{m} \mathbb{E}_{\epsilon}\bigg[\sup_{h \in \mathcal{H}_{C_{P},C_{1},...,C_{D},C_{Q}}^{d_{in}}}\sum_{i}\|h(a_{i})_{\cdot\cdot}\|_{p*,\infty}\bigg]  \tag{6} \label{eq:6}\\
&\leq \frac{N^{\frac{1}{p}}d_{u}}{m}\sum_{i} \sup_{h \in \mathcal{H}_{C_{P},C_{1},...,C_{D},C_{Q}}^{d_{in}}} \|h(a_{i})_{\cdot\cdot}\|_{p*,\infty}   \\
\end{flalign*}
Where we used norm inequality in \eqref{eq:6}.
\end{proof}

\noindent
\\
{\bf Theorem 1} {
Suppose $\mathcal{H}_{C_{P},C_{1},...,C_{D},C_{Q}}^{d_{in}}$ is a hypothesis class with constants $C_{P},C_{1},\dots,C_{D},C_{Q}$. Then, for samples $S=\{a_{i}\}_{i=1,\dots,m}$, we obtain the following inequality: 
\begin{equation*}
\begin{gathered}
\mathbb{E}_{\epsilon}\bigg[\frac{1}{m}\sup_{h \in \mathcal{H}_{C_{P},C_{1},...,C_{D},C_{Q}}^{d_{in}}}\sum_{i,\textbf{x},j}\epsilon_{i\textbf{x}j}h(a_{i})_{\textbf{x}j}\bigg] \\
\leq  L^{D}(NH)^{D\lfloor\frac{1}{p*}-\frac{1}{q} \rfloor_{+}}H^{\lfloor\frac{1}{p*}-\frac{1}{q}\rfloor_{+}}N^{\frac{1}{p}}{d_{u}}
C_{Q}C_{D}\dots C_{1}C_{P}\frac{1}{m}\sum_{i=1}^{m}\|{a_{i}}\|_{p*}.
\end{gathered}
\end{equation*}
} 
\begin{proof}
\begin{flalign*}
&\mathbb{E}_{\epsilon}\bigg[\frac{1}{m}\sup_{h \in \mathcal{H}_{C_{P},C_{1},...,C_{D},C_{Q}}^{d_{in}}}\sum_{i,\textbf{x},j}\epsilon_{i\textbf{x}j}h(a_{i})_{\textbf{x}j}\bigg] \\
&\leq N^{\frac{1}{p}}d_{u}\frac{1}{m}\sum_{i=1}^{m}\sup_{h \in \mathcal{H}_{C_{P},C_{1},...,C_{D},C_{Q}}}\|h(a_{i})_{\cdot\cdot}\|_{p*,\infty} \tag{Lemma 4} \\
&\leq L^{D}(NH)^{D\lfloor\frac{1}{p*}-\frac{1}{q} \rfloor_{+}}H^{\lfloor\frac{1}{p*}-\frac{1}{q}\rfloor_{+}}N^{\frac{1}{p}}d_{u}C_{Q}C_{D}\dots C_{1}C_{P}\frac{1}{m}\sum_{i=1}^{m}\|{a_{i}}\|_{p*}.  \tag{Lemma 3} \\
\end{flalign*}
\end{proof}
\noindent
{\bf Theorem 2 (FNO with a CNN layer)} {
Suppose ${\mathcal{H}_{CNN}}_{C_{P},C_{1},...,C_{D},C_{Q}}^{d_{in}}$ is a hypothesis class with constants $C_{P},C_{1},\dots,C_{D},C_{Q}$. Then, for samples $S=\{a_{i}\}_{i=1,\dots,m}$, we obtain the following inequality: 
\begin{equation*}
\begin{gathered}
\mathbb{E}_{\epsilon}\bigg[\frac{1}{m}\sup_{h \in {\mathcal{H}_{CNN}}_{C_{P},C_{1},...,C_{D},C_{Q}}^{d_{in}}}\sum_{i,\textbf{x},j}\epsilon_{i\textbf{x}j}h(a_{i})_{\textbf{x}j}\bigg] \\
\leq  L^{D}H^{(D+1)\lfloor\frac{1}{p*}-\frac{1}{q} \rfloor_{+}}N^{\frac{1}{p}}d_{u}
C_{Q}C_{D}\dots C_{1}C_{P}\frac{1}{m}\sum_{i=1}^{m}\|{a_{i}}\|_{p*}.
\end{gathered}
\end{equation*}

} 
\begin{proof} 
\begin{flalign*}
&\mathbb{E}_{\epsilon}\bigg[\frac{1}{m}\sup_{h \in {\mathcal{H}_{CNN}}_{C_{P},C_{1},...,C_{D},C_{Q}}^{d_{in}}}\sum_{i,\textbf{x},j}\epsilon_{i\textbf{x}j}h(a_{i})_{\textbf{x}j}\bigg] \\
&\leq N^{\frac{1}{p}}d_{u}\frac{1}{m}\sum_{i=1}^{m}\sup_{h \in {\mathcal{H}_{CNN}}_{C_{P},C_{1},...,C_{D},C_{Q}}}\|h(a_{i})_{\cdot\cdot}\|_{p*,\infty} \tag{Lemma 4} \\
&\leq L^{D}H^{(D+1)\lfloor\frac{1}{p*}-\frac{1}{q} \rfloor_{+}}N^{\frac{1}{p}}{d_{u}}C_{Q}C_{D}\dots C_{1}C_{P}\frac{1}{m}\sum_{i=1}^{m}\|{a_{i}}\|_{p*}.  \tag{Lemma 3'} \\
\end{flalign*}
\end{proof}
\noindent
{\bf Corollary 1} {For a constant $\gamma>0$, consider the hypothesis class $\mathcal{H}_{\gamma_{p,q}\leq\gamma}$, which is a collection of FNOs with $\gamma_{p,q}\leq\gamma$. For samples $S=\{a_{i}\}_{i=1,\dots,m}$, we obtain the following inequality: 
\begin{equation*}
\begin{gathered}
\mathbb{E}_{\epsilon}\bigg[\frac{1}{m}\sup_{h \in {\mathcal{H}_{\gamma_{p,q}\leq\gamma}}}\sum_{i,\textbf{x},j}\epsilon_{i\textbf{x}j}h(a_{i})_{\textbf{x}j}\bigg] \\
\leq \gamma_{}L^{D}(NH)^{D\lfloor\frac{1}{p*}-\frac{1}{q} \rfloor_{+}}H^{\lfloor\frac{1}{p*}-\frac{1}{q}\rfloor_{+}}N^{\frac{1}{p}}d_{u}\frac{1}{m}\sum_{i=1}^{m}\|{a_{i}}\|_{p*}.
\end{gathered}
\end{equation*}
For a given hypothesis class ${\mathcal{H}_{CNN}}_{\gamma_{p,q}\leq\gamma}$, similar to $\mathcal{H}_{\gamma_{p,q}\leq\gamma}$, and training samples $S=\{a_{i}\}_{i=1,\dots,m}$, we obtain the following inequality: 
\begin{equation*}
\begin{gathered}
\mathbb{E}_{\epsilon}\bigg[\frac{1}{m}\sup_{h \in {\mathcal{H}_{CNN}}_{\gamma_{p,q}\leq\gamma}}\sum_{i,\textbf{x},j}\epsilon_{i\textbf{x}j}h(a_{i})_{\textbf{x}j}\bigg] \\ \leq \gamma_{CNN}L^{D}H^{(D+1)\lfloor\frac{1}{p*}-\frac{1}{q} \rfloor_{+}}N^{\frac{1}{p}}d_{u}\frac{1}{m}\sum_{i=1}^{m}\|{a_{i}}\|_{p*}.
\end{gathered}
\end{equation*}
} 
\begin{proof}
As
\begin{flalign*}
&\mathcal{H}_{\gamma_{p,q}\leq\gamma} \subset \bigcup_{0\leq C_{P}C_{1}\dots C_{D}C_{Q}<\gamma}{\mathcal{H}_{C_{P},C_{1},...,C_{Q}}^{d_{in}} }.
\end{flalign*}
We have the following inequality:
\begin{flalign*}
&\mathbb{E}_{\epsilon}\bigg[\frac{1}{m}\sup_{h \in {\mathcal{H}_{\gamma_{p,q}\leq\gamma}}}\sum_{i,\textbf{x},j}\epsilon_{i\textbf{x}j}h(a_{i})_{\textbf{x}j}\bigg]  \\
&\leq\mathbb{E}_{\epsilon}\bigg[\frac{1}{m}\sup_{h\in \bigcup_{0\leq C_{P}C_{1}\dots C_{D}C_{Q}<\gamma}{\mathcal{H}_{C_{P},C_{1},...,C_{Q}}^{d_{in}} }}\sum_{i,\textbf{x},j}\epsilon_{i\textbf{x}j} h(a_{i})_{\textbf{x}j}\bigg]  \\
\end{flalign*}
Since the upper bound of $p*$-norm of models of hypothesis class in the above equation is the same as in Lemma 3, we can apply the same logic as in Theorem 1. So, we get the following inequality

\begin{flalign*}
&\mathbb{E}_{\epsilon}\bigg[\frac{1}{m}\sup_{h \in {\mathcal{H}_{\gamma_{p,q}\leq\gamma}}}\sum_{i,\textbf{x},j}\epsilon_{i\textbf{x}j}h(a_{i})_{\textbf{x}j}\bigg] \\
&\leq \gamma_{}L^{D}(NH)^{D\lfloor\frac{1}{p*}-\frac{1}{q} \rfloor_{+}}H^{\lfloor\frac{1}{p*}-\frac{1}{q}\rfloor_{+}}N^{\frac{1}{p}}d_{u}\frac{1}{m}\sum_{i=1}^{m}\|{a_{i}}\|_{p*}. 
\end{flalign*}
Similarly, based on the aforementioned proof, we obtain the inequality for the FNO with CNN layers.
\end{proof}
\noindent
Recall the following well-known theorem, which states statistical estimation of generalization error bound of given hypothesis class in terms of Rademacher complexity. This fundamental theorem can be found in (\cite{shalev:88}).\\

\noindent
{\bf Theorem 3 (Generalization error bounding based on the Rademacher complexity)} {For a given hypothesis class $\mathcal{H}$ and loss function $l:\mathcal{H}\times Z$ that satisfy the following case: for all $h \in \mathcal{H}$  and $z \in Z$, we obtain $\vert l(h,z)\vert \leq c$. Then, with a probability of at least $1-\delta$, for all $h\in \mathcal{H}$, we obtain
\begin{equation*}
\begin{gathered}
\mathbb{E}_{\mathcal{D}}[l(h,z)]-\mathbb{E}_{S}[l(h,z)] \leq 2\mathcal{R}_{m}(l \circ\mathcal{ H})+c\sqrt{\frac{2\log{4/\delta}}{m}}.
\end{gathered}
\end{equation*}
where $\mathcal{D}$ is probability distribution on $Z$ and $S$ is a training dataset sampled from $\mathcal{D}$ i.i.d.

}

\noindent
\\
Before considering generalization bound of FNO, we choose distribution $\mathcal{D}$ on $\mathbb{R}^{N\times d_{a}}\times\mathbb{R}^{N\times d_{u}}$ to have a compact support. So that $\vert l(h,z)\vert\leq c$ condition in Theorem 3 holds. Then, using the aforementioned theorem 3 and our corollary 1, we obtain the following estimation of the generalization error bound: \\
\\
\noindent
\textbf{Theorem 4 (Generalization error bound for FNO)} {For the training dataset $S=\{(a_{i},u_{i})\}_{i=1,\dots,m}$ which is sampled from probability distribution $\mathcal{D}$ i.i.d., and for hypothesis class $\mathcal{H}_{\gamma_{p,q}\leq\gamma}$, let $h^{\star}$ be the ERM minimizer of $L_{S}$ and suppose $\|h(a)-u\|_{2} \leq \epsilon^{2}$ for all $(a,u)\sim\mathcal{D}$, $h\in\mathcal{H}_{\gamma_{p,q}\leq\gamma}$. Then, with a probability of at least $1-\delta$, we obtain the following inequality:
\begin{equation*}
\begin{gathered}
L_{\mathcal{D}}(h^{\star})-L_{S}(h^{\star}) \\ 
\leq 4\sqrt{2}\epsilon \gamma_{}L^{D}(NH)^{D\lfloor\frac{1}{p*}-\frac{1}{q} \rfloor_{+}}H^{\lfloor\frac{1}{p*}-\frac{1}{q}\rfloor_{+}}  
N^{\frac{1}{p}}d_{u}\frac{1}{m}\sum_{i=1}^{m}\|{a_{i}}\|_{p*}+\epsilon^{2}\sqrt{\frac{2\log{4/\delta}}{m}}.
\end{gathered}
\end{equation*}  
Similarly for hypothesis class of FNOs with CNN layers, for dataset $S$, and for hypothesis class ${\mathcal{H}_{CNN}}_{\gamma_{p,q}\leq\gamma}$, let $h^{\star}_{CNN}$ be the ERM minimizer of $L_{S}$ and suppose $\|h(a)-u\|_{2} \leq \epsilon^{2}$ for all $(a,u)\sim\mathcal{D}$, $h\in{\mathcal{H}_{CNN}}_{\gamma_{p,q}\leq\gamma}$. Then, with a probability of at least $1-\delta$, we obtain the following inequality:
\begin{equation*}
\begin{gathered}
L_{\mathcal{D}}(h^{\star}_{CNN})-L_{S}(h^{\star}_{CNN}) \\ 
\leq 4\sqrt{2}\epsilon \gamma_{CNN}L^{D}H^{(D+1)\lfloor\frac{1}{p*}-\frac{1}{q} \rfloor_{+}}N^{\frac{1}{p}}d_{u}\frac{1}{m}\sum_{i=1}^{m}\|{a_{i}}\|_{p*} +\epsilon^{2}\sqrt{\frac{2\log{4/\delta}}{m}}.
\end{gathered}
\end{equation*}  
}
\begin{proof} We just need to calculate $\mathcal{R}_{m}(l \circ\mathcal{ H})$ term in Theorem 3.
\begin{flalign*}
&\mathcal{R}_{m}(l \circ\mathcal{H}_{\gamma_{p,q}\leq\gamma(\mathcal{N}_{FNO} )}) \leq 2\sqrt{2}\epsilon \mathbb{E}_{\epsilon}\bigg[\frac{1}{m}\sup_{h \in {\mathcal{H}_{\gamma_{p,q}\leq\gamma}}}\sum_{i,\textbf{x},j}\epsilon_{i\textbf{x}j}h(a_{i})_{\textbf{x}j}\bigg]  \tag{Lemma 2} \\[5pt]
&\leq  2\sqrt{2}\epsilon\gamma_{}L^{D}(NH)^{D\lfloor\frac{1}{p*}-\frac{1}{q} \rfloor_{+}}H^{\lfloor\frac{1}{p*}-\frac{1}{q}\rfloor_{+}} N^{\frac{1}{p}}d_{u}\frac{1}{m}\sum_{i=1}^{m}\|{a_{i}}\|_{p*}.   \tag{Corollary 1} \label{Coro}    
\end{flalign*}
Similarly, based on the aforementioned proof, we obtain the inequality for the FNO with CNN layers.
\end{proof}
\noindent
Now, if the capacity of FNO model $h$ is $\gamma$, it is included in the hypothesis class $\mathcal{H}_{\gamma_{p,q}\leq\gamma}$. Since inequalities in theorem 4 hold for all hypotheses in class, we have the following posterior estimate of FNO.\\

\noindent
{\textbf{Corollary 2 (Posterior estimation of generalization error and expected error)}}{
For given architecture parameters $N, H, d_{u}, d_{a}, L$, and training samples $\{(a_{i},u_{i})\}_{i=1,\dots,m}$ with $\|a_{i}\|_{p*} \leq B$ for all $i$. Suppose $h$ is trained FNO (Fourier layers with FCN or CNN) such that $\|h(a)-u\|_{2} \leq \epsilon^{2}$ for all training samples. Then with the confidence of at least $1-\delta$, we have the following estimates.
\begin{equation*}
\begin{gathered}
L_{\mathcal{D}}(h)-L_{S}(h) \\ 
\leq 4\sqrt{2}\epsilon L^{D}(NH)^{D\lfloor\frac{1}{p*}-\frac{1}{q} \rfloor_{+}}H^{\lfloor\frac{1}{p*}-\frac{1}{q}\rfloor_{+}}  
N^{\frac{1}{p}}d_{u}\gamma_{p,q}(h)B
+\epsilon^{2}\sqrt{\frac{2\log{4/\delta}}{m}}.
\end{gathered}
\end{equation*}
\begin{equation*}
\begin{gathered}
\Longrightarrow L_{\mathcal{D}}(h) 
\leq 4\sqrt{2}\epsilon L^{D}(NH)^{D\lfloor\frac{1}{p*}-\frac{1}{q} \rfloor_{+}}H^{\lfloor\frac{1}{p*}-\frac{1}{q}\rfloor_{+}}  
N^{\frac{1}{p}}d_{u}\gamma_{p,q}(h)B
+\epsilon^{2}\bigg(1+\sqrt{\frac{2\log{4/\delta}}{m}}\bigg).
\end{gathered}
\end{equation*}
And for FNOs with CNN,
\begin{equation*}
\begin{gathered}
L_{\mathcal{D}}(h_{CNN})-L_{S}(h_{CNN}) \\\leq 4\sqrt{2}\epsilon L^{D}H^{(D+1)\lfloor\frac{1}{p*}-\frac{1}{q} \rfloor_{+}}N^{\frac{1}{p}}d_{u}{\gamma_{CNN}}_{p,q}(h_{CNN})B
+\epsilon^{2}\sqrt{\frac{2\log{4/\delta}}{m}}.
\end{gathered}
\end{equation*}
\begin{equation*}
\begin{gathered}
\Longrightarrow L_{\mathcal{D}}(h_{CNN}) 
\leq 4\sqrt{2}\epsilon L^{D}H^{(D+1)\lfloor\frac{1}{p*}-\frac{1}{q} \rfloor_{+}}N^{\frac{1}{p}}d_{u}{\gamma_{CNN}}_{p,q}(h_{CNN})B
+\epsilon^{2}\bigg(1+\sqrt{\frac{2\log{4/\delta}}{m}}\bigg).
\end{gathered}
\end{equation*}
}

\section{Experiments}\label{sec4}
In this section, we validate our results based on experiments. First,
we investigate the correlation between our capacity and the empirical generalization errors for various capacities of $p$ and $q$. Hereafter, we check the dependency of
generalization errors on the model architecture by varying $k_{max}$.  \\
\noindent
{\bf Data specification} {For our experiment, we synthesize the dataset based on the following Burgers equation problem:
\begin{equation*}
\begin{gathered}
u_{t}=-uu_{x}+0.01uu_{xx}
\end{gathered}
\end{equation*}
The domain of the problem is a circle, and we uniformly discretize the domain by $N=1024$. As described in Section 2, each data point is a pair of functions. In our experiment, the input function is an initial condition, and the target function is a solution to the aforementioned equation at $t=0.5$. Each input function is generated from Gaussian random fields with covariance $k(x,y)=e^{-\frac{(x-y)^2}{(0.05)^2}}$. The training dataset comprises 800 pairs of functions, and the test dataset comprises 200 pairs of functions (both generated independently).
}

\noindent
{\bf Correlation for various capacities of $p$ and $q$} {We checked the correlation between the generalization error and capacities. Each point in Figure \ref{pq} represents a trained model for the randomly chosen hyperparameters. The architecture of the models in our experiment is organized as follows: 2-depth Fourier layers, linear layers without activation for projection, and lifting layers. The width is fixed at 64. The weight decay for each training session is randomly chosen from 0, 20*1e-3, 40*1e-3, 60*1e-3, and 80*1e-3; $k_{max}$ is randomly chosen from the values of 8, 12, 16, and 20; the size of the kernel is randomly chosen from 1, 3, 5, and 7 for 100 iterations.
\begin{figure}[htp!]
\centering
\includegraphics[height=5.7cm]{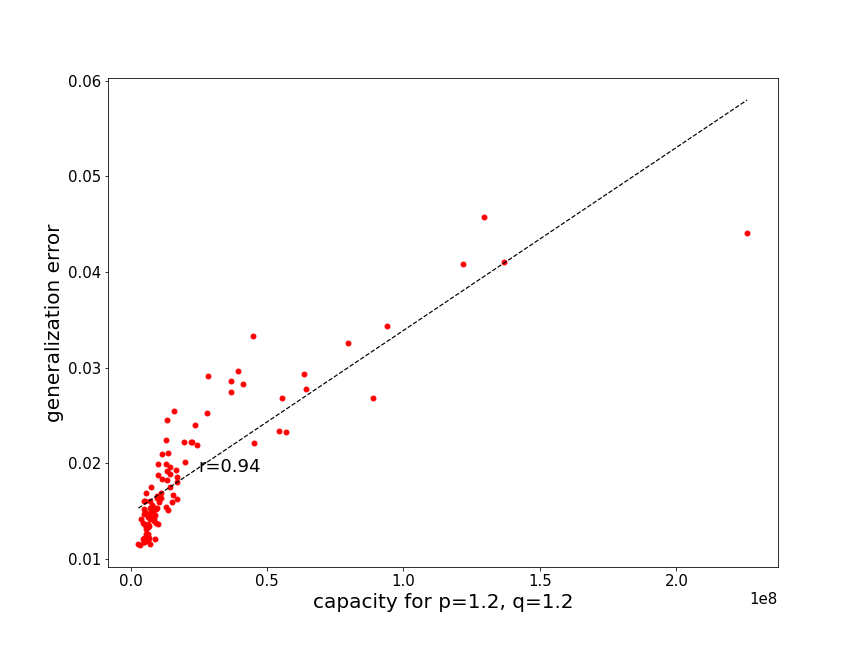}
\caption{Scatter plot of generalization error vs capacity for $p=1.2,q=1.2$}\label{pq}
\end{figure}

\begin{table}[H]
\centering
\begin{tabular}{|c|c|c|c|c|c|c|}
\noalign{\smallskip}\noalign{\smallskip}\hline
& $p=1$ & $p=1.2$ & $p=1.6$ & $p=2$ & $p=4$ & $p=\infty$\\
\hline
$q=1$ & 0.8757 & 0.9137 & 0.7794 & 0.7595 & 0.7542 &0.7285\\
\hline
$q=1.2$ & 0.8395 & \textbf{0.9358} & 0.8007 & 0.7635 & 0.7526 &0.7265\\
\hline
$q=1.6$ & 0.8127 & 0.9007 & 0.8476 & 0.7750 & 0.7495&0.7231\\
\hline
$q=2$ & 0.8037 & 0.8720 & 0.8815 & 0.7860 & 0.7466 &0.7204\\
\hline
$q=4$ & 0.7919 & 0.8417 & 0.9084 & 0.7938 & 0.7322&0.7112\\
\hline
$q=\infty$ & 0.7555 & 0.8229 & 0.8859 & 0.7765 & 0.7235&0.7219\\
\hline
\end{tabular}
\caption{Correlation between empirical generalization error and capacities of various $p$ and $q$ for trained models with randomly chosen hyperparameters} \label{t1}
\end{table}
\noindent
Table \ref{t1} lists the correlations for various values of $p$ and $q$. As observed, the correlation tends to decrease when the $p$ and $q$ values increase. This is because, as $p$ increases, the $p$-norm loses information about elements other than the highest one. Thus, the information of the model itself is lost in a capacity defined by high values of $p$ and $q$. However, instead of losing information about model weights, as $p$ goes to $\infty$, $p*$ goes to 1; thus, the terms about kernel size and $k_{max}$ are more affecting to capacity as $p$ increases. Therefore, we assume that capacities of high $p$ contain more information about the model architecture. To prove our arguments, we conduct experiments in which $k_{max}$ vary and other hyperparameters are fixed. First, to show that capacities of low $p$ and $q$ have more information about the model's weights than its architecture, we trained three types of models that have negligible differences in $k_{max}$. Second, to show that capacities of high $p$ and $q$ are more related to the model architecture, we trained three kinds of models that have considerable differences in $k_{max}$. For each experiment, we trained the models 30 times for each $k_{max}$ setting, i.e., 14, 16, and 18 for the left column of Figure \ref{fig:subfigures4} and 10, 30, and 50 for the right column. Hyperparameters other than $k_{max}$ are fixed: the kernel size of the CNN layer is 1, the width is 64, and the depth of the Fourier layers is 2. As evident in the left column of Figure \ref{fig:subfigures4}, models with small gaps in $k_{max}$ lose the correlation between the generalization gap and capacity as the $p$ and $q$ values increase. However, as shown in the right column of Figure \ref{fig:subfigures4}, the highest correlation between capacity and generalization error is obtained for higher $p$ and $q$ values compared with those of the left column case. The correlation is maintained at ~0.89 for the $p,q =\infty$ case.
} 

\begin{figure}[H]
    \begin{center}
     
        \includegraphics[width=0.40\textwidth]{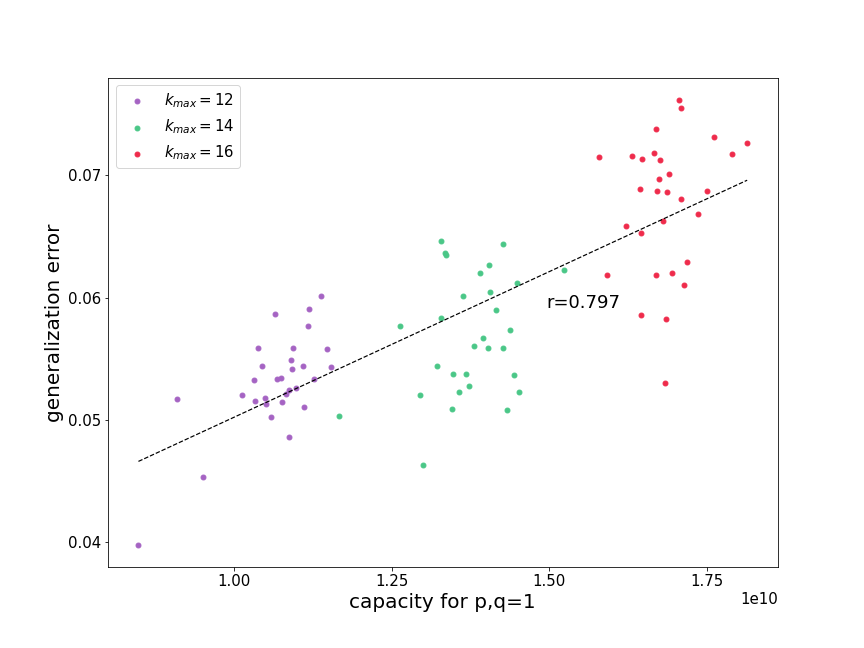}
        \includegraphics[width=0.40\textwidth]{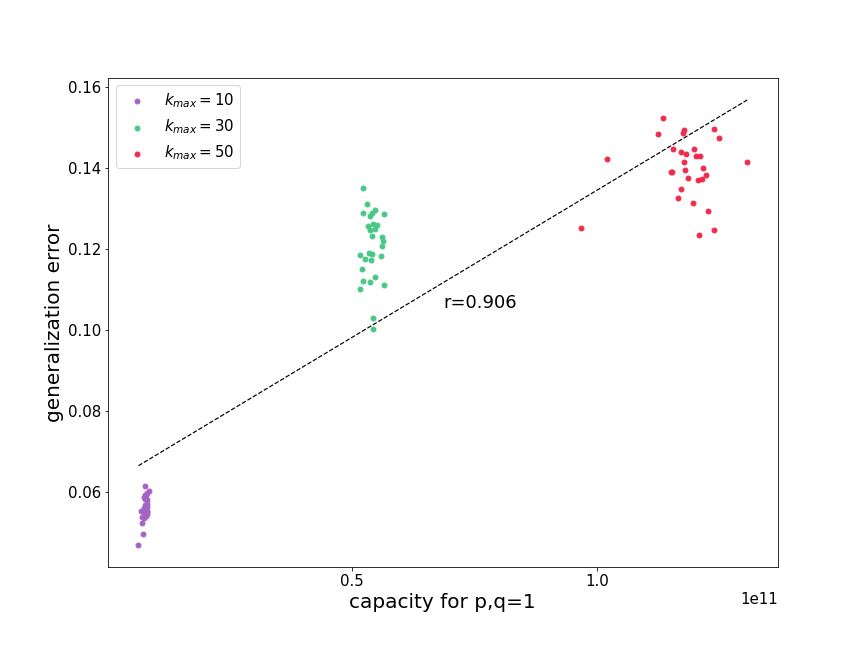}
        \\[\smallskipamount]
        \includegraphics[width=0.40\textwidth]{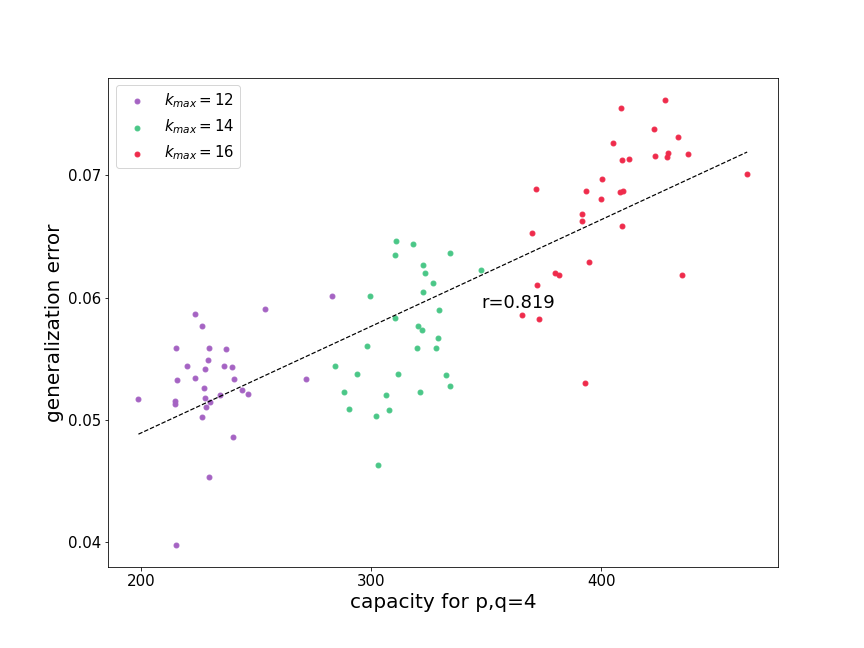}
        \includegraphics[width=0.40\textwidth]{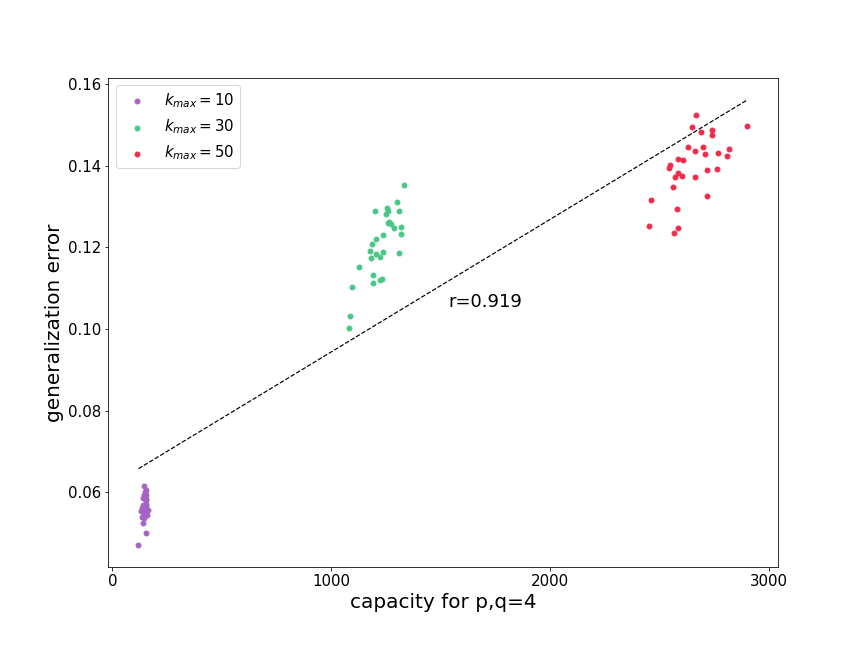}
        \\[\smallskipamount]
        \includegraphics[width=0.40\textwidth]{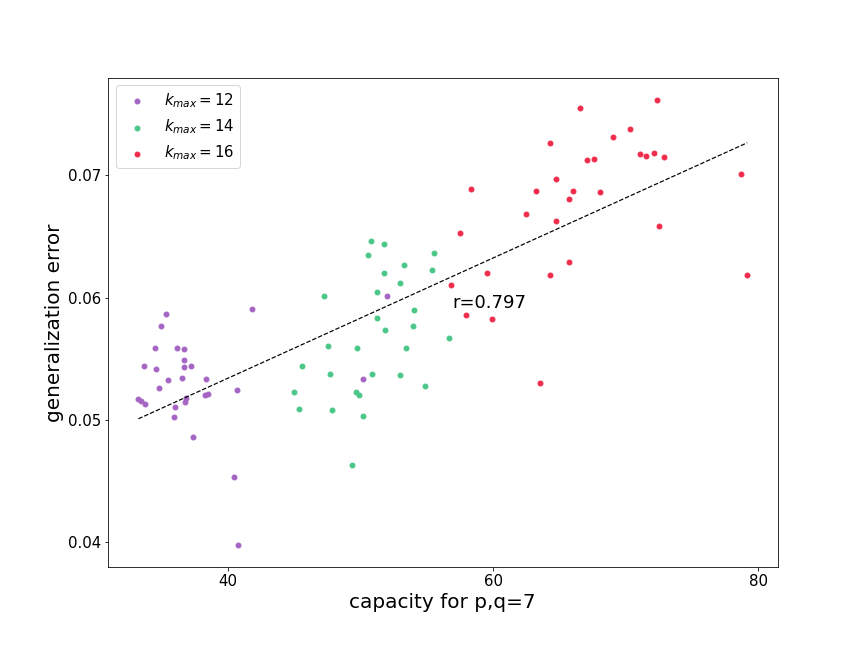}
        \includegraphics[width=0.40\textwidth]{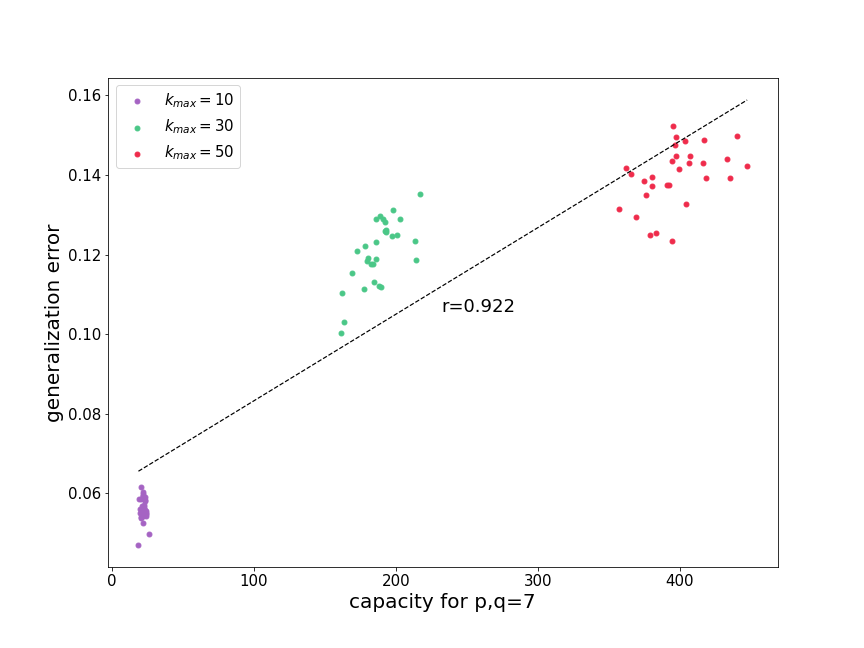}
        \\[\smallskipamount]
        \includegraphics[width=0.40\textwidth]{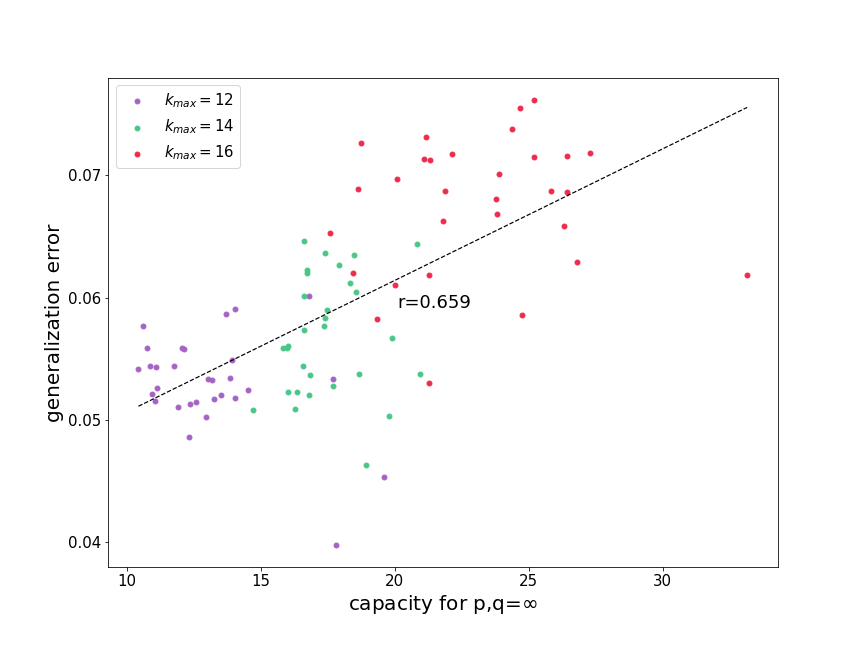}
        \includegraphics[width=0.40\textwidth]{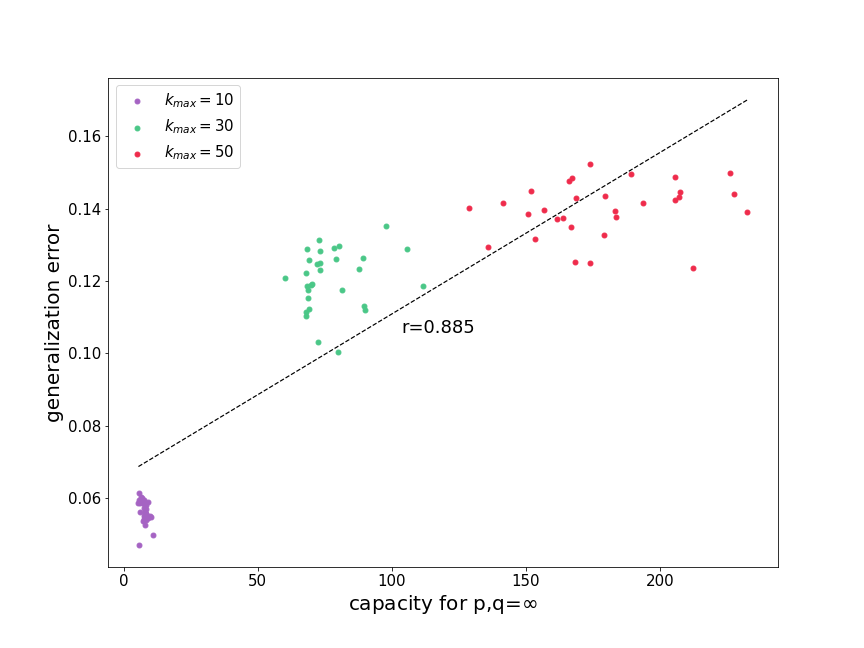}
      
    \end{center}
    \caption{%
        \textbf{Left:} Scatter plot, correlation and linear regression between generalization error and capacities of various p and q for 30 trained models for each $k_{max}=14, 16, 18$;
        \textbf{Right:} Scatter plot, correlation and linear regression between generalization error and capacities of various p and q for 30 trained models for each $k_{max}=10, 30, 50$.
     }%
     
   \label{fig:subfigures4}
   
\end{figure}

\noindent
{\bf Dependency on $k_{max}$} {Next, we examined the dependency of the generalization error on the model architecture. In the experiments, the hyperparameters other than $k_{max}$ were fixed. We considered two cases: Fourier layers at depths of 1 and 2. 
Low $k_{max}$ implies that the dynamics of learning are unpredictable and chaotic (\cite{Sele:22}); thus, we did not consider models with too small values of $k_{max}$. We varied $k_{max}$ from 13 to 39 in two intervals. For a detailed analysis, we removed the CNN layer parts in the Fourier layers, such that 
the generalization error was proportional to weight norm $R_{i}$. To check the influence of the size of $k_{max}$, we divided the generalization error by $R_{i}$. As the defined capacity is correlated to the generalization error, it is expected that 
divided generalization error is correlated to $\sqrt[\leftroot{-1}\uproot{4}p*]{k_{max}}$ at a depth of 1 and $\sqrt[\leftroot{-1}\uproot{4}\frac{p*}{2}]{k_{max}}$ at a depth of 2. As listed in Tables \ref{t2} and \ref{t3}, the generalization error divided by the norms is correlated to $\sqrt[\leftroot{-1}\uproot{4}p*]{k_{max}}$ and $\sqrt[\leftroot{-1}\uproot{4}\frac{p*}{2}]{k_{max}}$.
Hence, we could verify that for low capacities of $p$ and $q$, the tendency of correlation was low and the desired dependency on $k_{max}$ was quite unclear. As $p$ and $q$ increased, the correlation increased. Then, as $p$ and $q$ increased further, the correlation slightly dropped . Based on these data, we can conclude that capacities with higher $p$ and $q$ have more information about the model architecture($k_{max}$), and capacities of very high $p$ and $q$ may cause loss of specific information about each model; thus, the correlation drops down. Figure \ref{fig:subfigures5} shows a scatter plot and regression for a few cases of our experiments. The generalization error dependency on $k_{max}$ is more convex at the depth of 2. Based on our definition of capacity, the exponent of the $k_{max}$ term is proportional to the depth of Fourier layers. Therefore, the increased convexity illustrated on the right side of the figures qualitatively validates our results.
\begin{table}[H]
\centering
\begin{tabular}{|c|c|c|c|c|c|c|c|}
\noalign{\smallskip}\noalign{\smallskip}\hline
& $p=2$ & $p=2.5$ & $p=4$ & $p=8$ & $p=20$ & $p=\infty$\\
\hline
$q=1$ & 0.6913 & 0.8199 & 0.8928 & 0.9062 & 0.8855 &0.8647\\
\hline
$q=2$ & 0.7210 & 0.8386 & 0.9029 & \textbf{0.9129} & 0.8921 &0.8699\\
\hline
$q=4$ & 0.7302 & 0.8389 & 0.8990 & 0.9064 & 0.8868 &0.8629\\
\hline
$q=8$ & 0.7041 & 0.8133 & 0.8797 & 0.8872 & 0.8649 &0.8328\\
\hline
$q=\infty$ & 0.6561 & 0.7620 & 0.8454 & 0.8573 & 0.8231 &0.7741\\
\hline
\end{tabular}
\caption{Correlation between empirical generalization error divided by weight norm of Fourier layers and $\sqrt[\leftroot{-1}\uproot{4}p*]{k_{max}}$ for FNO with depth 1 of Fourier layers} \label{t2}
\end{table}
\begin{table}[H]
\centering
\begin{tabular}{|c|c|c|c|c|c|c|c|}
\noalign{\smallskip}\noalign{\smallskip}\hline
& $p=2$ & $p=4$ & $p=8$ & $p=12$ & $p=20$ & $p=\infty$\\
\hline
$q=1$ & -0.4145 & 0.8722 & 0.9319 & 0.9387 & 0.9385 &0.9322\\
\hline
$q=2$ & -0.4027 & 0.8882 & 0.9396 & 0.9439 & 0.9436 &0.9365\\
\hline
$q=4$ & -0.3386 & 0.9063 & 0.9484 & 0.9506 & 0.9485 &0.9397\\
\hline
$q=8$ & -0.1041 & 0.9129 & \textbf{0.9508} & 0.9504 & 0.9448 &0.9319\\
\hline
$q=\infty$ & 0.3099 & 0.8821 & 0.9207 & 0.9162 & 0.9045 &0.8834\\
\hline
\end{tabular}
\caption{Correlation between empirical generalization error divided by weight norm of Fourier layers and $\sqrt[\leftroot{-1}\uproot{4}\frac{p*}{2}]{k_{max}}$ for FNO with depth 2 of Fourier layers} \label{t3}
\end{table}

\begin{figure}[H]

     \begin{center}

        \includegraphics[width=0.40\textwidth]{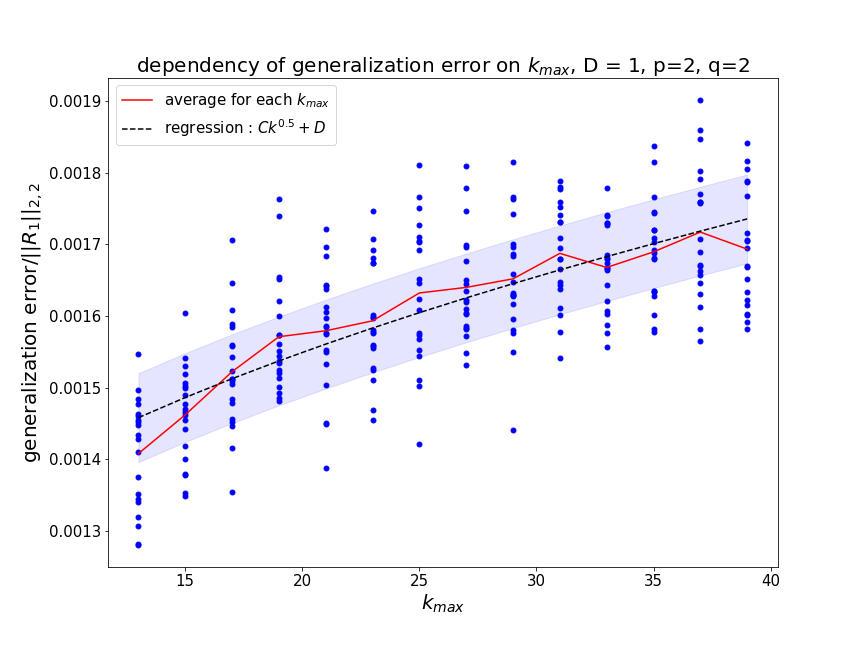}
        \includegraphics[width=0.40\textwidth]{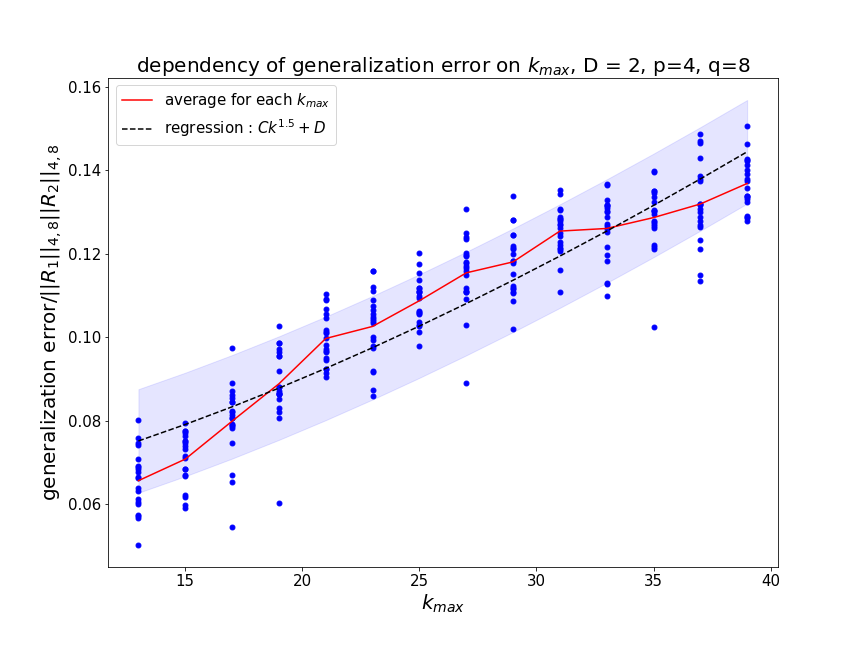}
        \\[\smallskipamount]
        \includegraphics[width=0.40\textwidth]{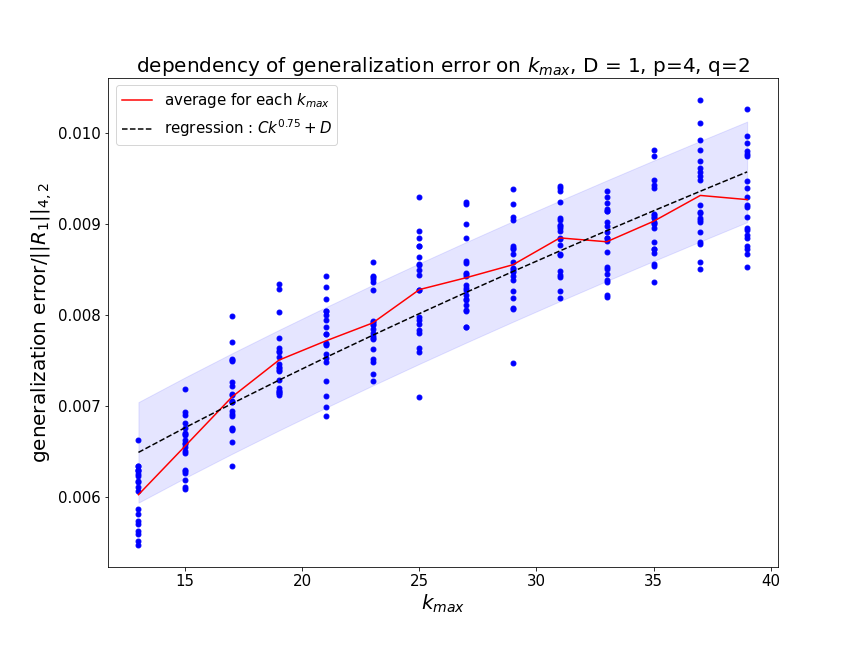}
        \includegraphics[width=0.40\textwidth]{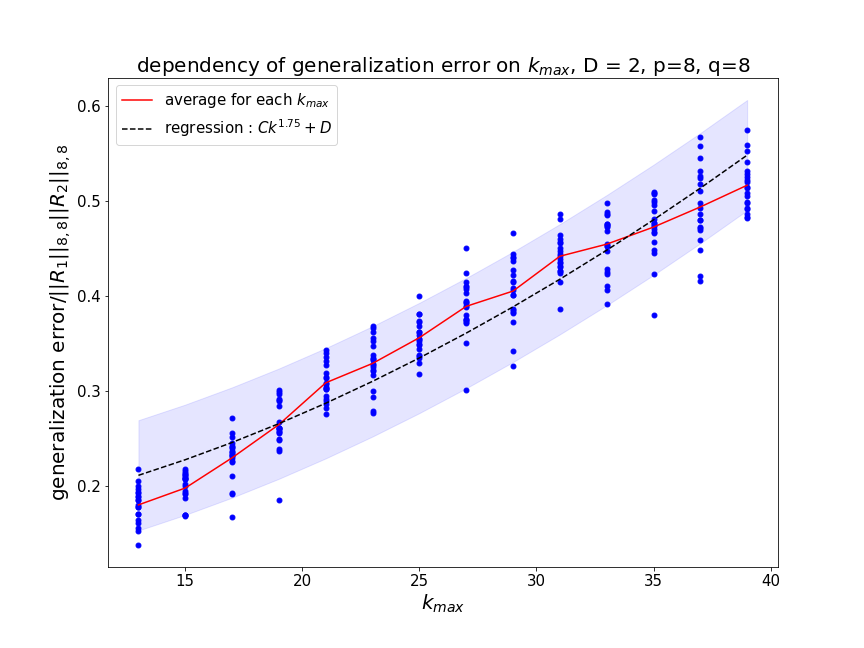}
        \\[\smallskipamount]
        \includegraphics[width=0.40\textwidth]{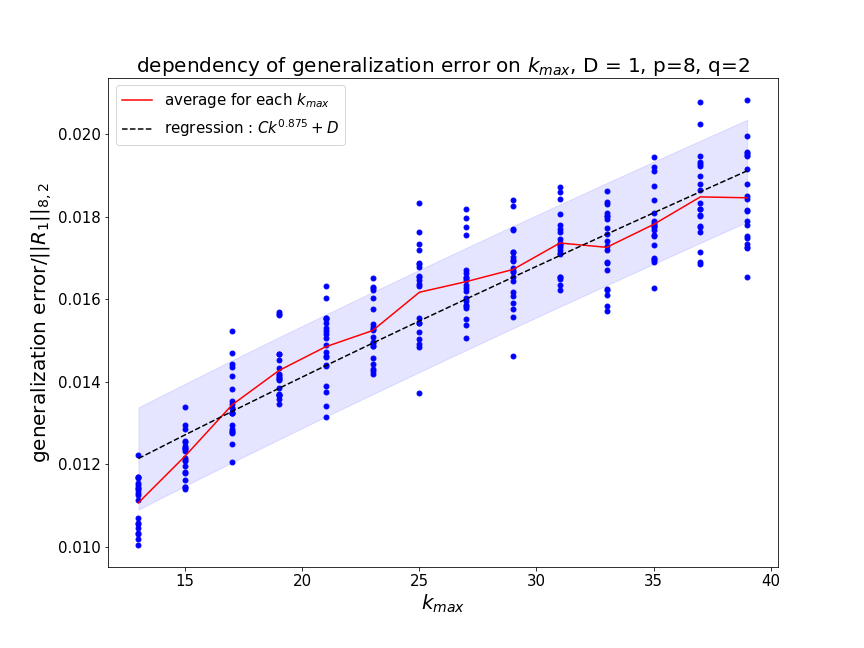}
        \includegraphics[width=0.40\textwidth]{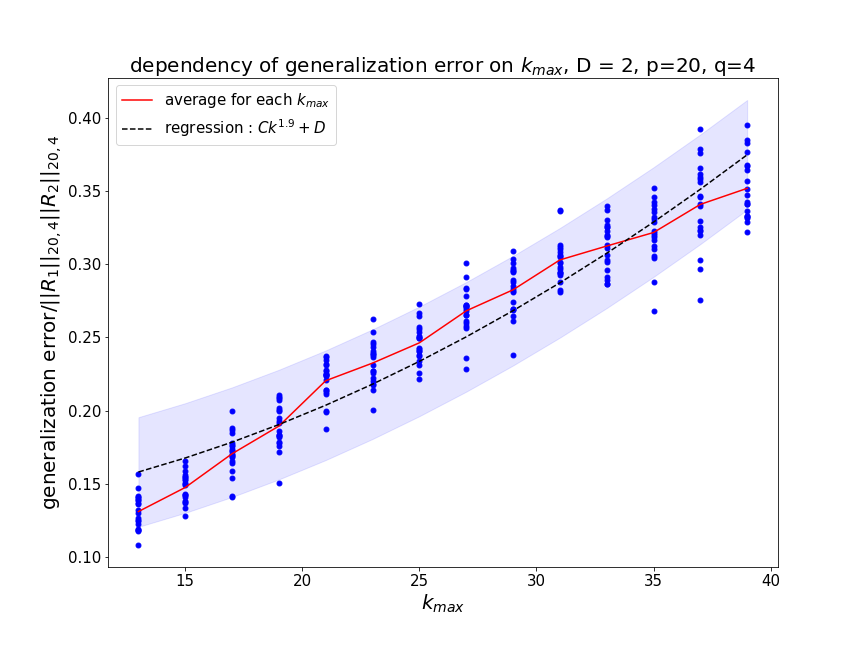}
        \\[\smallskipamount]
        \includegraphics[width=0.40\textwidth]{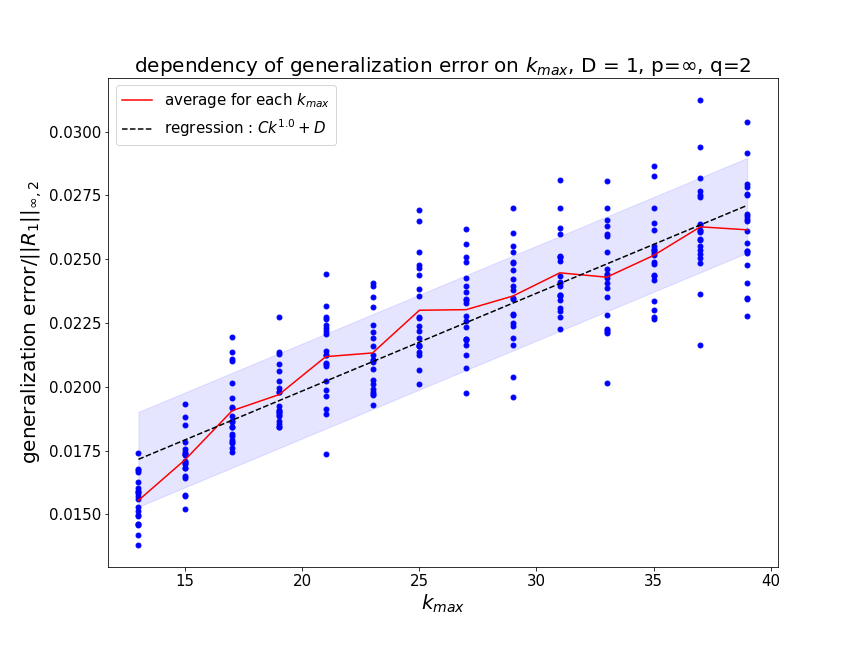}
        \includegraphics[width=0.40\textwidth]{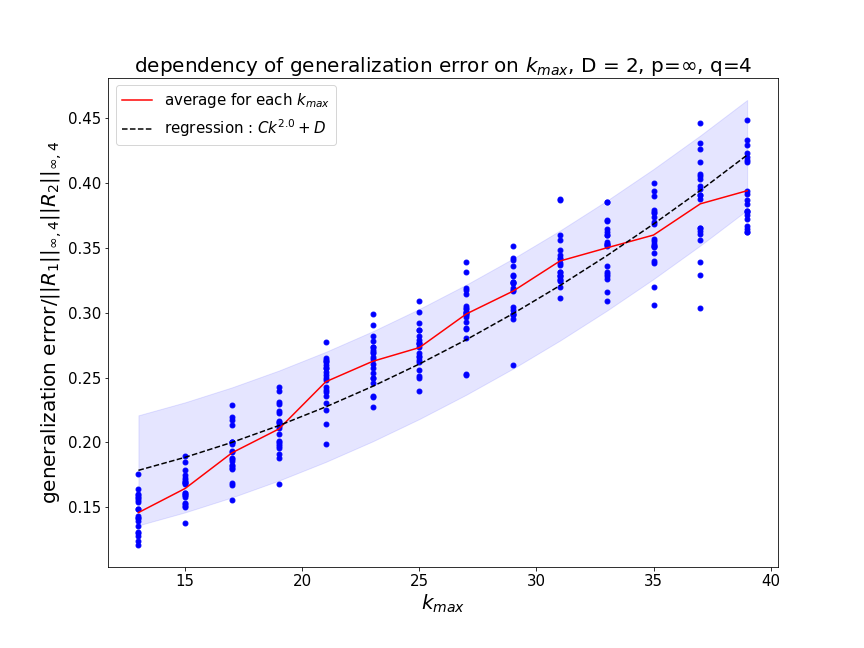}
      
   \end{center}
    \caption{%
        \textbf{Left:} Scatter plot and regression between generalization error divided by norms of fourier layers and $\sqrt[\leftroot{-1}\uproot{4}p*]{k_{max}}$ for various $p, q$ where depth of fourier layer is 1;
        \textbf{Right:} Scatter plot and regression between generalization error divided by norms of fourier layers and $\sqrt[\leftroot{-1}\uproot{4}\frac{p*}{2}]{k_{max}}$ for various $p, q$ where depth of fourier layer is 2.
     }%
   \label{fig:subfigures5}
\end{figure}

\section{Conclusion}\label{sec5}
We investigated the bounding Rademacher complexity of an FNO and defined its capacity, which depends on the model architecture and the group norm of the weights. Although several results  already exist with regard to the bounding Rademacher complexity of various types of neural networks, the FNO possesses tensor weights that rank higher than two. Therefore, our study may be helpful for other NNs that contain higher-rank tensors. We validated our results through experiments. Based on these experiments, we gained insights into the impact of $p$ and $q$ values and the information about model weights and architecture stored in terms of capacities. Various neural operators have been developed, including FNO and DeepONet; however, the analysis of PAC learning to these neural operators has not been performed in detail. Thus, this study may serve as a guide for such analysis. 
Herein, the original FNO by (\cite{Li:21}) was implemented with the GeLU activation function, which contains various parameters. In this study, we assumed the activation function to be fixed. For a general model containing parameterized activation, e.g., GeLU, we need to modify our analysis. Although the Rademacher complexity contains information about datasets, the bounding of our results lacks specific dependency on each problem. As we experimented with various PDE problems, the performance of the FNO varied for each problem. Therefore, we need to extend the complexities to include information about the datasets. 

\appendix

\vskip 0.2in
\bibliography{FNO}

\begin{thebibliography}{30}
\providecommand{\natexlab}[1]{#1}
\providecommand{\url}[1]{\texttt{#1}}
\expandafter\ifx\csname urlstyle\endcsname\relax
  \providecommand{\doi}[1]{doi: #1}\else
  \providecommand{\doi}{doi: \begingroup \urlstyle{rm}\Url}\fi

\bibitem[Awasthi et~al.(2020)Awasthi, Frank, and Mohri]{Awasthi:20}
P.~Awasthi, N.~Frank, and M.~Mohri.
\newblock On the rademacher complexity of linear hypothesis sets.
\newblock \emph{arXiv}, arXiv:2007.11045, 2020.

\bibitem[Bartlett et~al.(2021)Bartlett, Foster, and Telgarsky]{Bart:21}
P.L. Bartlett, D.J. Foster, and M.~Telgarsky.
\newblock Spectrally-normalized margin bounds for neural networks.
\newblock \emph{Advances in Neural Information Processing Systems}, 2021.

\bibitem[Cai et~al.(2021)Cai, Chen, and Liu]{Cai:21}
Z.~Cai, J.~Chen, and M.~Liu.
\newblock Least-squares relu neural network (lsnn) method for linear
  advection-reaction equation.
\newblock \emph{Journal of Computational Physics}, 443:\penalty0 686--707,
  2021.

\bibitem[G.~Gupta and Bogdan(2021)]{Gupta:21}
X.~Xiao G.~Gupta and P.~Bogdan.
\newblock Multiwavelet-based operator learning for differential equations.
\newblock \emph{Advances in Neural Information Processing Systems}, 2021.

\bibitem[Gopalani et~al.(2022)Gopalani, Karmakar, and Mukherjee]{Gopalani:22}
P.~Gopalani, S.~Karmakar, and A.~Mukherjee.
\newblock Capacity bounds for the deeponet method of solving differential
  equations.
\newblock \emph{arXiv}, arXiv:2205.11359, 2022.

\bibitem[Hao et~al.(2019)Hao, Zhanfeng, Zhouwang, and Xiao]{Hao:19}
C.~Hao, M.~Zhanfeng, Y.~Zhouwang, and W.~Xiao.
\newblock Theoretical investigation of generalization bound for residual
  networks.
\newblock \emph{International Joint Conferences on Artificial Intelligence
  Organization}, pages 2081--2087, 2019.

\bibitem[Jakubovitz et~al.(2019)Jakubovitz, Giryes, and Rodrigues]{Jaku:19}
D.~Jakubovitz, R.~Giryes, and M.R.D. Rodrigues.
\newblock \emph{Generalization Error in Deep Learning}.
\newblock Birkhäuser Cham, 2019.

\bibitem[Kovachki et~al.(2021{\natexlab{a}})Kovachki, Lanthaler, and
  Mishra]{Kovachki:21}
N.~Kovachki, S.~Lanthaler, and S.~Mishra.
\newblock On universal approximation and error bounds for fourier neural
  operators.
\newblock \emph{Journal of Machine Learning Research}, 22\penalty0 (290),
  2021{\natexlab{a}}.

\bibitem[Kovachki et~al.(2021{\natexlab{b}})Kovachki, Li, Liu, Azizzadenesheli,
  Bhattacharya, Stuart, and Anandkumar]{Kovachki:212}
N.~Kovachki, Z.~Li, B.~Liu, K.~Azizzadenesheli, K.~Bhattacharya, A.~Stuart, and
  A.~Anandkumar.
\newblock Neural operator: Learning maps between function spaces.
\newblock \emph{arXiv}, arXiv:2108.08481, 2021{\natexlab{b}}.

\bibitem[Lei et~al.(2019)Lei, Dogan, Zhou, and Kloft]{Lei:19}
Y.~Lei, Ü. Dogan, D.~Zhou, and M.~Kloft.
\newblock Data-dependent generalization bounds for multi-class classification.
\newblock \emph{IEEE Transactions on Information Theory}, 65\penalty0
  (5):\penalty0 2995--3021, 2019.

\bibitem[Li et~al.(2020)Li, Kovachki, Azizzadenesheli, Liu, Bhattacharya,
  Stuart, and Anandkumar]{Li:20}
Z.~Li, N.~Kovachki, K.~Azizzadenesheli, B.~Liu, K.~Bhattacharya, A.~Stuart, and
  A.~Anandkumar.
\newblock Neural operator: Graph kernel network for partial differential
  equations.
\newblock \emph{ICLR 2020 Workshop ODE/PDE+DL}, 2020.

\bibitem[Li et~al.(2021)Li, Kovachki, Azizzadenesheli, Liu, Bhattacharya,
  Stuart, and Anandkumar]{Li:21}
Z.~Li, N.~Kovachki, K.~Azizzadenesheli, B.~Liu, K.~Bhattacharya, A.~Stuart, and
  A.~Anandkumar.
\newblock Fourier neural operator for parametric partial differential
  equations.
\newblock \emph{ICLR 2021}, 2021.

\bibitem[Liang et~al.(2019)Liang, Poggio, Rakhlin, and Stokes]{Liang:19}
T.~Liang, T.~Poggio, A.~Rakhlin, and J.~Stokes.
\newblock Fisher-rao metric, geometry, and complexity of neural networks.
\newblock \emph{Proceedings of the Twenty-Second International Conference on
  Artificial Intelligence and Statistics}, 89:\penalty0 888--896, 2019.

\bibitem[Long and Sedghi(2020)]{Long:20}
P.M. Long and H.~Sedghi.
\newblock Generalization bounds for deep convolutional neural networks.
\newblock \emph{ICLR 2020}, 2020.

\bibitem[Lu et~al.(2021)Lu, Jin, Pang, Zhang, and Karniadakis]{Lu:21}
L.~Lu, P.~Jin, G.~Pang, Z.~Zhang, and G.E. Karniadakis.
\newblock Learning nonlinear operators via deeponet based on the universal
  approximation theorem of operators.
\newblock \emph{Nature Machine Intelligence}, 3\penalty0 (3):\penalty0
  218--229, 2021.

\bibitem[Lv(2021)]{Lv:21}
S.~Lv.
\newblock Generalization bounds for graph convolutional neural networks via
  rademacher complexity.
\newblock \emph{arXiv}, arXiv:2102.10234, 2021.

\bibitem[Maurer(2016)]{Maurer:16}
A.~Maurer.
\newblock A vector-contraction inequality for rademacher complexities.
\newblock \emph{arXiv}, arXiv:1605.00251, 2016.

\bibitem[Minshuo et~al.(2020)Minshuo, Xingguo, and Tuo]{Minshuo:20}
C.~Minshuo, L.~Xingguo, and Z.~Tuo.
\newblock On generalization bounds of a family of recurrent neural networks.
\newblock \emph{Proceedings of Machine Learning Research}, 108:\penalty0
  1233--1243, 2020.

\bibitem[Neyshabur et~al.(2015)Neyshabur, Tomioka, and Srebro]{Neysh:15}
B.~Neyshabur, R.~Tomioka, and N.~Srebro.
\newblock Norm-based capacity control in neural networks.
\newblock \emph{Proceedings of Machine Learning Research}, 40:\penalty0
  1376--1401, 2015.

\bibitem[Pathak et~al.(2022)Pathak, subramanian, Harrington, Raja,
  Chattopadhyay, Mardani, Kurth, Hall, Li, Azizzadenesheli, Hassanzadeh,
  Kashinath, and Anandkumar]{Pathak:22}
J.~Pathak, S~subramanian, P.~Harrington, S.~Raja, A.~Chattopadhyay, M.~Mardani,
  T.~Kurth, D.~Hall, Z.~Li, K.~Azizzadenesheli, P.~Hassanzadeh, K.~Kashinath,
  and A.~Anandkumar.
\newblock Fourcastnet: A global data-driven high-resolution weather model using
  adaptive fourier neural operators.
\newblock \emph{arXiv}, arXiv:2202.11214, 2022.

\bibitem[Petzka et~al.(2021)Petzka, Kamp, Adilova, Sminchisescu, and
  Boley]{Petzka:21}
H.~Petzka, M.~Kamp, L.~Adilova, C.~Sminchisescu, and M.~Boley.
\newblock Relative flatness and generalization.
\newblock \emph{Advances in Neural Information Processing Systems}, 2021.

\bibitem[Raissi et~al.(2019)Raissi, Perdikaris, and Karniadakis]{Raissi:19}
M.~Raissi, P.~Perdikaris, and G.E. Karniadakis.
\newblock Physics-informed neural networks: A deep learning framework for
  solving forward and inverse problems involving nonlinear partial differential
  equations.
\newblock \emph{Journal of Computational Physics}, 378:\penalty0 686--707,
  2019.

\bibitem[Seleznova and Kutyniok(2021)]{Sele:22}
M.~Seleznova and G.~Kutyniok.
\newblock Analyzing finite neural networks: Can we trust neural tangent kernel
  theory?
\newblock \emph{Proceedings of Machine Learning Research}, 145:\penalty0
  847--867, 2021.

\bibitem[Shalev-Shwartz and Ben-David(2014)]{shalev:88}
Shai Shalev-Shwartz and Shai Ben-David.
\newblock \emph{Understanding Machine Learning: From Theory to Algorithms}.
\newblock Cambridge University Press, San MateoShaftesbury Road, Cambridge,
  2014.

\bibitem[Sontag(1998)]{Sontag:98}
E.D. Sontag.
\newblock Vc dimension of neural networks.
\newblock 1998.

\bibitem[Valiant(1984)]{Valiant:84}
L.G. Valiant.
\newblock A theory of the learnable.
\newblock \emph{Communications of the ACM}, 27\penalty0 (11):\penalty0
  1134--1142, 1984.

\bibitem[Vapnik(1999)]{Vapnik:21}
V.N. Vapnik.
\newblock An overview of statistical learning theory.
\newblock \emph{IEEE Transactions on Neural Networks}, 10\penalty0
  (5):\penalty0 988--999, 1999.

\bibitem[Weinan and Yu(2018)]{Weinan:18}
E.~Weinan and B.~Yu.
\newblock The deep ritz method: A deep learning-based numerical algorithm for
  solving variational problems.
\newblock \emph{Communications in Mathematics and Statistics}, 6\penalty0
  (1):\penalty0 1--12, 2018.

\bibitem[Weinan et~al.(2020)Weinan, Chao, and Qingcan]{Weinan:20}
E.~Weinan, M.~Chao, and W.~Qingcan.
\newblock Rademacher complexity and the generalization error of residual
  networks.
\newblock \emph{Communications in Mathematical Sciences}, 18\penalty0
  (6):\penalty0 1755--1774, 2020.

\bibitem[Wen et~al.(2022)Wen, Li, Azizzadenesheli, Anandkumar, and
  Benson]{Wen:22}
G.~Wen, Z.~Li, K.~Azizzadenesheli, A.~Anandkumar, and S.M. Benson.
\newblock U-fno—an enhanced fourier neural operator-based deep-learning model
  for multiphase flow.
\newblock \emph{Advances in Water Resources}, 163, 2022.

\end{thebibliography}
\nocite{Vapnik:21,Gopalani:22,Lei:19,Long:20,Lv:21,Jaku:19,Minshuo:20,Hao:19,Kovachki:212,Wen:22,Awasthi:20}

\end{document}